\documentclass{article}
\pdfoutput=1  
\PassOptionsToPackage{numbers,sort&compress}{natbib}
\usepackage[preprint]{neurips_2026}
\usepackage[utf8]{inputenc}
\usepackage[T1]{fontenc}
\usepackage{hyperref}
\usepackage{url}
\usepackage{booktabs}
\usepackage{amsfonts}
\usepackage{amsmath}
\usepackage{amssymb}
\usepackage{nicefrac}
\usepackage{microtype}
\usepackage{xcolor}
\usepackage{graphicx}
\usepackage{algorithm}
\usepackage{algorithmic}
\usepackage{multirow}
\usepackage{subcaption}
\usepackage{wrapfig}
\usepackage[most]{tcolorbox}
\usepackage{tcolorbox}
\usepackage{tikz}
\usetikzlibrary{positioning,arrows.meta,shapes.geometric,fit,backgrounds,calc}
\usepackage{pgfplots}
\pgfplotsset{compat=1.18}
\usepackage{amsmath,amssymb}
\usepackage{pifont}
\usepackage{tikz}
\usetikzlibrary{arrows.meta,calc,positioning,decorations.pathreplacing,shapes.geometric}

\title{On-Policy Distillation with Best-of-$N$ Teacher Rollout Selection}

\author{%
  Ke Zhang\thanks{Work done during an internship at TikTok.} \\
  Johns Hopkins University \\
  \And
  Yunjie Tian \\
  TikTok \\
  \And
  DongDi Zhao \\
  TikTok \\
  \AND
  Yijiang Li \\
  University of California, San Diego\\
  \And
  Yuanye Liu \\
  Fudan University \\
  \\
  \And
  Vishal M.\ Patel \\
  Johns Hopkins University \\
  \And
  Di Fu\thanks{Corresponding author.} \\
  TikTok\\
}

\begin{document}
\maketitle

\begin{abstract}
On-policy distillation (OPD), which supervises a student on its own sampled trajectories, has emerged as a data-efficient post-training method for improving reasoning while avoiding the reward dependence of reinforcement learning and the catastrophic forgetting often observed in standard supervised fine-tuning.
However, standard OPD typically computes teacher supervision under noisy student-generated contexts and often relies on a single stochastic teacher rollout per prompt. As a result, the supervision signal can be high-variance: the sampled teacher trajectory can be incorrect, uninformative, or poorly matched to the student's current reasoning behavior.
To address this limitation, we propose \textbf{BRTS}, a \textbf{B}est-of-$N$ \textbf{R}ollout \textbf{T}eacher \textbf{S}election framework for on-policy distillation. BRTS augments standard student-context OPD with a teacher-context supervision branch constructed from the curated teacher trajectory. Rather than distilling from the first sampled teacher rollout, BRTS samples a small pool of teacher trajectories and selects the auxiliary trajectory using a simple priority rule: correctness first, student alignment second.
When multiple correct teacher trajectories are available, BRTS chooses the one most aligned with the student's current behavior; when unconditioned teacher samples fail on harder prompts, it invokes a ground-truth-conditioned recovery step to elicit a natural derivation. 
The selected trajectory is then used to provide reliable teacher-context supervision inside the OPD loop, augmented with an auxiliary loss on the teacher trajectory.
Experiments on AIME 2024, AIME 2025, and AMC 2023 show that BRTS improves over standard OPD on challenging reasoning benchmarks, with the largest gains on harder datasets.
Our code is available at \url{https://github.com/BWGZK-keke/BRTS}.
\end{abstract}

\begin{figure}[t]
\centering

\definecolor{tcblue}{HTML}{2C3E6E}
\definecolor{tcblueText}{HTML}{283858}
\definecolor{wrred}{HTML}{B33A3A}
\definecolor{wrredText}{HTML}{FD93B7}
\definecolor{trgreen}{RGB}{94,169,248}
\definecolor{contour}{HTML}{A8B8D5}
\definecolor{contourpink}{HTML}{D8B8C5}
\definecolor{panelfr}{HTML}{9AA3AE}
\definecolor{flowtxt}{HTML}{1A1A1A}
\definecolor{captxt}{HTML}{2A2A2A}
\definecolor{labeltxt}{HTML}{1A1A1A}

\definecolor{corrblue}{RGB}{94,169,248}

\definecolor{replaceclr}{HTML}{C77A2E}

\definecolor{strongBlue}{HTML}{C2D0EA}
\definecolor{coreBlue}{HTML}{F2F5FB}
\definecolor{midBlue}{HTML}{F2F2FB}
\definecolor{outerPink}{HTML}{FFF7FB}

\tikzset{
  every node/.style    = {font=\rmfamily},
  panel/.style         = {draw=panelfr, line width=0.6pt, rounded corners=2pt},
  contour/.style       = {draw=contour, line width=0.5pt, dash pattern=on 2pt off 1.6pt},
  contourout/.style    = {draw=contourpink, line width=0.5pt, dash pattern=on 2pt off 1.6pt},
  arr/.style           = {-{Stealth[length=5.6pt,width=4.4pt]}, line width=1.0pt},
  flowarr/.style       = {arr, color=black!82},
  redarr/.style        = {arr, color=wrred, dashed, dash pattern=on 2.8pt off 1.6pt},
  greenarr/.style      = {arr, color=trgreen, line width=1.15pt},
  replacearr/.style    = {-{Stealth[length=5.0pt,width=4.0pt]}, color=replaceclr, line width=0.9pt, dash pattern=on 2.4pt off 1.4pt},
  sampleline/.style    = {draw=black!68, line width=0.75pt, dash pattern=on 3.0pt off 1.8pt},
  correctedline/.style = {draw=trgreen, line width=0.85pt, dash pattern=on 3.0pt off 1.8pt},
  node-teacher/.style  = {circle, draw=white, line width=0.6pt, fill=tcblue, inner sep=0pt, minimum size=5.8pt},
  node-corrected/.style= {circle, draw=trgreen, line width=1.1pt, fill=white, inner sep=0pt, minimum size=5.8pt},
  node-student/.style  = {circle, draw=black!80, line width=1.0pt, fill=white, inner sep=0pt, minimum size=5.8pt},
  pcap/.style          = {font=\rmfamily\small, align=center, text=captxt},
  mlabelbig/.style     = {font=\rmfamily\small, text=labeltxt},
  smalllabel/.style    = {font=\rmfamily\scriptsize, text=captxt},
  flowlabel/.style     = {font=\rmfamily\footnotesize, text=flowtxt},
  reglabel/.style      = {font=\rmfamily\small, text=labeltxt},
  regsub/.style        = {font=\rmfamily\footnotesize, text=labeltxt},
  kllabel/.style       = {font=\rmfamily\scriptsize, text=black!78},
}

\newcommand{\PW}{7.4}
\newcommand{\PH}{5.6}

\newcommand{\drawManifold}[2]{%
  \begin{scope}
    \clip (0,0) rectangle (\PW,\PH);
    \fill[outerPink] (0,0) rectangle (\PW,\PH);
    \fill[midBlue]   (#1,#2) ellipse (2.85cm and 2.35cm);
    \fill[midBlue]  (#1,#2) ellipse (2.15cm and 1.80cm);
    \fill[midBlue] (#1,#2) ellipse (1.0cm and 0.85cm);

    \foreach \rx/\ry in {1.0/0.85, 1.55/1.30, 2.15/1.80}
      { \draw[contour] (#1,#2) ellipse (\rx cm and \ry cm); }
    \foreach \rx/\ry in {2.85/2.35, 3.55/2.95}
      { \draw[contourout] (#1,#2) ellipse (\rx cm and \ry cm); }
  \end{scope}
}

\newcommand{\drawRegionLabels}{%
  \node[regsub,   text=corrblue, align=center] at (1.85,5.1) {True Reasoning Manifold};
  \node[regsub,   text=wrredText, align=center] at (5.55,5.1) {Errors / Hallucinations};
}

\resizebox{\textwidth}{!}{%
\begin{tikzpicture}[x=1cm, y=1cm]

\begin{scope}[shift={(0,0)}]
  \drawManifold{1.85}{2.6}
  \begin{scope}
    \clip (0.05,0.05) rectangle (\PW-0.05,\PH-0.05);
    \drawRegionLabels

    \coordinate (CenterA) at (1.85,2.60);
    \coordinate (TeachA)  at (4.30,2.30);

    \coordinate (TrueAOne)    at (3.00,3.45);
    \coordinate (TrueATwo)    at (2.80,2.78);
    \coordinate (TrueAThree)  at (1.68,1.78);
    \coordinate (WrongAOne)   at (5.55,3.20);
    \coordinate (WrongATwo)   at (5.95,2.10);
    \coordinate (WrongAThree) at (5.30,1.30);

    \draw[sampleline] (TeachA) -- (TrueAOne);
    \draw[sampleline] (TeachA) -- (TrueATwo);
    \draw[sampleline] (TeachA) -- (TrueAThree);
    \draw[sampleline] (TeachA) -- (WrongAOne);
    \draw[sampleline] (TeachA) -- (WrongATwo);
    \draw[sampleline] (TeachA) -- (WrongAThree);

    \node[font=\rmfamily\normalsize, text=trgreen] at (TrueAOne) {\checkmark};
    \node[font=\rmfamily\normalsize, text=trgreen] at (TrueATwo) {\checkmark};
    \node[font=\rmfamily\normalsize, text=trgreen] at (TrueAThree) {\checkmark};
    \node[font=\rmfamily\normalsize, text=wrred] at (WrongAOne) {$\times$};
    \node[font=\rmfamily\normalsize, text=wrred] at (WrongATwo) {$\times$};
    \node[font=\rmfamily\normalsize, text=wrred] at (WrongAThree) {$\times$};

    \node[star, star points=5, star point ratio=2.4, fill=corrblue, draw=white,
          line width=0.3pt, inner sep=0.6pt, minimum size=10pt] at (CenterA) {};
    \node[flowlabel, anchor=south] at (1.85,2.65) {$y^{*}$ ground truth};

    \node[node-teacher] (TA) at (TeachA) {};
    \node[mlabelbig, anchor=west] at (4.42,2.42) {teacher $\pi_{T}$};

    \node[node-student] (SA) at (5.95,0.85) {};
    \node[mlabelbig, anchor=west] at (5.90,0.70) {student $\pi_{S}$};

    \draw[flowarr] (SA) .. controls (5.70,1.55) and (5.00,1.95) .. (TA);
    \node[flowlabel] at (6,1.55) {Distill};
  \end{scope}
\end{scope}
\node[pcap] at (\PW/2,-0.42) {(a) OPD: errors propagate};

\begin{scope}[shift={(\PW+0.6,0)}]
  \drawManifold{1.85}{2.6}
  \begin{scope}
    \clip (0.05,0.05) rectangle (\PW-0.05,\PH-0.05);
    \drawRegionLabels

    \coordinate (CenterB) at (1.85,2.60);
    \coordinate (TeachB)  at (4.30,2.30);

    \coordinate (TrueBOne)     at (3.00,3.45);
    \coordinate (TrueBTwo)     at (2.80,2.78);
    \coordinate (TrueBThree)   at (1.68,1.78);
    \coordinate (TrueBFour)   at (3,1.78);
    \coordinate (WrongBOne)    at (5.55,3.20);
    \coordinate (WrongBTwo)    at (5.95,2.10);
    \coordinate (WrongBThree)  at (5.30,1.30);

    \coordinate (CorrB)        at (3.50,3.92);
    \coordinate (CorrCloseB)   at (2.5,2.2);

    \draw[sampleline] (TeachB) -- (TrueBOne);
    \draw[sampleline] (TeachB) -- (TrueBTwo);
    \draw[sampleline] (TeachB) -- (TrueBThree);
    \draw[sampleline] (TeachB) -- (WrongBOne);
    \draw[sampleline] (TeachB) -- (WrongBTwo);
    \draw[sampleline] (TeachB) -- (WrongBThree);
    \draw[sampleline] (TeachB) -- (CorrCloseB);

    \node[font=\rmfamily\normalsize, text=trgreen] at (TrueBOne) {\checkmark};
    \node[font=\rmfamily\normalsize, text=trgreen] at (TrueBTwo) {\checkmark};
    \node[font=\rmfamily\normalsize, text=black!45] at (TrueBThree) {\checkmark};
    \node[font=\rmfamily\normalsize, text=wrred] at (WrongBOne) {$\times$};
    \node[font=\rmfamily\normalsize, text=wrred] at (WrongBTwo) {$\times$};
    \node[font=\rmfamily\normalsize, text=wrred] at (WrongBThree) {$\times$};

    \node[star, star points=5, star point ratio=2.4, fill=corrblue, draw=white,
          line width=0.3pt, inner sep=0.6pt, minimum size=10pt] at (CenterB) {};
    \node[flowlabel, anchor=south] at (1.85,2.65) {$y^{*}$ ground truth};

    \node[node-teacher] (TB) at (TeachB) {};
    \node[mlabelbig, anchor=west] at (4.45,2.46) {teacher $\pi_{T}$};

    \node[node-corrected] (TC) at (3.3,2.4) {};
    \node[mlabelbig, anchor=north] at (3.1,1.9) {$\pi_{T}\mid y^{*}$};
    \node[flowlabel, anchor=north] at (3.1,1.5) {corrected};

    \node[font=\rmfamily\normalsize, text=trgreen] at (CorrB) {\checkmark};
    \node[font=\rmfamily\normalsize, text=trgreen] at (CorrCloseB) {\checkmark};

    \draw[correctedline] (TC) -- (TrueBOne);
    \draw[correctedline] (TC) -- (TrueBTwo);
    \draw[correctedline] (TC) -- (CorrB);
    \draw[correctedline] (TC) -- (CorrCloseB);

    \draw[-{Stealth[length=5.0pt,width=4.0pt]}, color=black!50, line width=0.9pt]
      ($(TrueBThree) + (0.20,0.10)$) -- ($(CorrCloseB) + (-0.20,-0.10)$);
    \node[flowlabel, text=black!55, anchor=south, rotate=27] at (2.05,2.00) {resample};

    \node[node-student] (SB) at (5.95,0.85) {};
    \node[mlabelbig, anchor=west] at (5.90,0.70) {student $\pi_{S}$};

\draw[redarr]
  (WrongBOne)
  .. controls (5.10,3.55) and (4.45,4.18) ..
  (CorrB);

\node[flowlabel, text=wrred] at (4.45,4.3)
  {condition on $y^{*}$};

\draw[greenarr] (SB) .. controls (5.10,0.55) and (3.95,0.85) .. (TC);
\node[flowlabel, text=trgreen] at (4.60,0.30) {Distill toward $\pi_{T}|y^{*}$};
  \end{scope}
\end{scope}
\node[pcap] at (\PW+0.6+\PW/2,-0.42) {(b) BRTS: errors corrected, distant samples aligned};

\end{tikzpicture}%
}

\caption{Conceptual comparison of OPD and BRTS. 
(a) Classical OPD may propagate unreliable teacher signals when teacher trajectories are incorrect or poorly matched to the student. 
(b) BRTS constructs correctness- and alignment-aware teacher-context supervision by selecting or recovering a reliable teacher trajectory before distillation.}
\label{fig:teaser}
\end{figure}
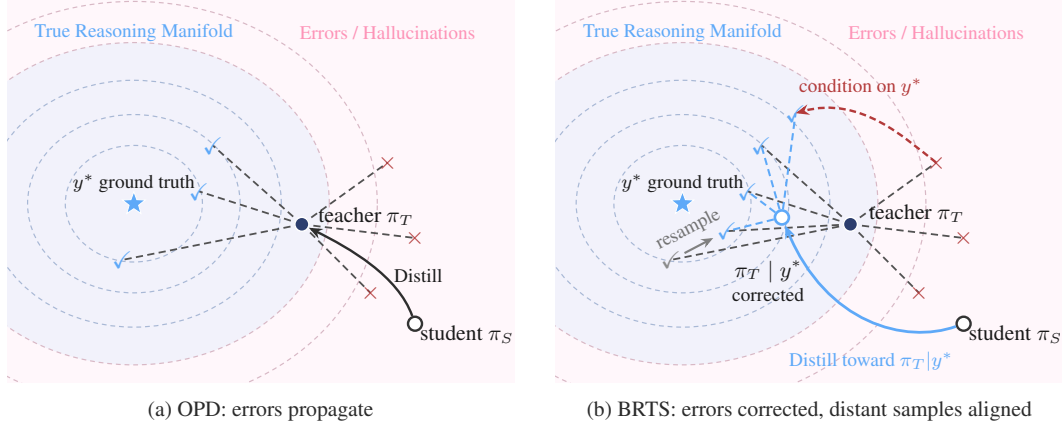

\section{Introduction}
\label{sec:intro}

On-policy distillation (OPD) has rapidly become a standard tool in large language model (LLM) post-training~\cite{agarwal2024gkd,gu2023minillm,lu2025opd,yang2025qwen3,xiao2026mimo,ye2026onpolicycontext,ye2026gad,jang2026stable,jin2026entropy,ko2026scaling}. The recipe is appealingly simple: have the student generate rollouts from its own policy, and use the teacher's per-token log-probabilities on those rollouts as a dense supervision signal. 
Compared with supervised fine-tuning (SFT) on teacher-generated text or classical sequence-level distillation~\cite{hinton2015kd,kim2016seqkd,sanh2019distilbert,jiao2020tinybert,wang2020minilm,kdsurvey}, OPD is less susceptible to exposure bias because supervision is defined on exactly the states the student visits at inference time~\cite{bengio2015scheduled,ross2011dagger,xu2020error}. Industry post-training pipelines~\cite{yang2025qwen3,xiao2026mimo,zeng2026glm5} have adopted OPD as a competitive complement to SFT and outcome-reward reinforcement learning~\cite{shao2024DeepSeekmath,yu2025dapo,guo2025DeepSeekr1,chu2025sft,shenfeld2025rlrazor,chen2025retaining}, with reported comparable gains at a fraction of the RL compute cost~\cite{lu2025opd}.
Despite this appeal, recent work has begun to unpack when OPD works and why~\cite{li2026rethinking,fu2026revisiting,kim2026does}. 
Two conditions appear to govern success: the student and teacher should share compatible reasoning patterns, and the teacher be able to produce correct, naturally derived solutions beyond the student’s current exploration range, thereby transferring new capabilities on difficult tasks.
This observation is also consistent with recent findings that small models can struggle to imitate strong reasoners when the teacher's reasoning style is poorly matched to the student~\cite{fu2023specializing,li2025small,guha2025openthoughts}.

Standard OPD reduces exposure bias by evaluating teacher feedback on student-generated trajectories, but this also means that supervision is computed on prefixes produced by an imperfect student~\cite{bengio2015scheduled,ross2011dagger,agarwal2024gkd,lu2025opd}. On difficult prompts, these prefixes can drift into noisy reasoning states, where local teacher feedback is less informative, and the student may never observe a complete correct solution path~\cite{li2026rethinking,fu2026revisiting,li2025small}. 
Practical OPD-style pipelines therefore often use teacher-side signals, such as reference rollouts, solvability estimates, metadata, or filtering rules, to identify prompts or trajectories that provide useful supervision~\cite{lu2025opd,xiao2026mimo,yang2026genonpolicy,li2026rethinking}. 
However, the teacher is also stochastic: for a fixed prompt, its sampled trajectories can differ substantially in correctness, reasoning style, and proximity to the student, especially on hard prompts~\cite{wang2023selfconsistency,zelikman2022star,lightman2023lets,singh2024beyond}.
Thus, a single teacher sample can provide a high-variance estimate of the teacher’s competence and alignment for a given prompt. When this sample is incorrect or poorly matched to the student, it can produce noisy guidance ( Figure~\ref{fig:teaser} (a)).
To address this limitation, we introduce \textbf{BRTS}, a 
\textbf{C}orrectness- and \textbf{A}lignment-aware 
\textbf{T}eacher \textbf{S}upervision framework for on-policy distillation.
As illustrated in Figure~\ref{fig:teaser}(b), BRTS first identifies a reliable teacher trajectory and then distills it as an additional teacher-context supervision signal, rather than learning from an arbitrary teacher rollout.
This complements standard student-context distillation: while OPD corrects the states visited by the student, BRTS exposes the student to complete and reliable teacher reasoning paths.
The teacher-context branch is both correctness-aware and alignment-aware. Correctness prevents erroneous teacher samples from being amplified, while alignment keeps the selected trajectory close to the student's current reasoning distribution.
For prompts where ordinary teacher sampling fails, BRTS uses a ground-truth-guided recovery step to obtain a teacher derivation, keeping the auxiliary branch active on hard examples where reliable supervision is most needed.
%
%
In this way, BRTS turns stochastic teacher rollouts into structured supervision inside the OPD loop.
Experiments on three public datasets show that BRTS improves over standard OPD, with larger gains on harder prompts where reliable teacher trajectories are sparse.




Our contributions are summarized as follows:
\begin{itemize}
    \item We identify a high-variance supervision issue in OPD: standard student-context supervision can be limited by noisy student prefixes and stochastic teacher rollouts, making the teacher signal unreliable on hard prompts.
    
    \item We introduce \textit{BRTS}, a Best-of-$N$ rollout teacher selection framework that provides a lightweight mechanism for selecting and distilling reliable teacher trajectories.
    With an answer-hint fallback for failed unconditioned samples, BRTS distills the selected trajectory through a ground truth-guided teacher-context objective.
    
    \item BRTS significantly improves over standard OPD on the hard benchmarks. We also provide diagnostic analyses showing how teacher candidate quality, recovery coverage, and teacher backbone choice affect the resulting OPD performance.
\end{itemize}

\section{Related Work}

\paragraph{From off-policy to on-policy distillation.}
Knowledge distillation transfers behavior and capabilities from a stronger teacher to a smaller or weaker student~\cite{hinton2015kd,kim2016seqkd,sanh2019distilbert,jiao2020tinybert,wang2020minilm,kdsurvey}. For autoregressive language models, standard off-policy distillation on teacher-generated text introduces a distribution mismatch: the student is trained under teacher-induced contexts but must condition on its own prefixes at inference time~\cite{bengio2015scheduled,ross2011dagger,xu2020error}. Optimizing supervised objectives on fixed teacher outputs may further over-constrain the student to trajectories outside its reachable distribution, increasing learning difficulty~\cite{zhang2025towards,xu2024speculative} and causing catastrophic forgetting~\cite{luo2025empirical,shenfeld2025rl}. On-policy distillation (OPD) alleviates this by letting the student generate trajectories and using the teacher to supervise the states the student actually visits. MiniLLM~\cite{gu2023minillm} and GKD~\cite{agarwal2024gkd} formulate this with reverse-KL and related divergences, while recent systems such as Qwen3~\cite{yang2025qwen3}, MiMo~\cite{xiao2026mimo}, and GLM-5~\cite{zeng2026glm5} combine distillation with supervised fine-tuning and outcome-reward RL~\cite{shao2024DeepSeekmath,yu2025dapo,guo2025DeepSeekr1}. Recent variants extend OPD to black-box, contextual, entropy-aware, and reward-extrapolated settings~\cite{lu2025opd,ye2026gad,ye2026onpolicycontext,jin2026entropy,yang2026genonpolicy}.

\paragraph{Understanding and improving on-policy distillation.}
Recent work has begun to analyze when OPD succeeds or fails. \cite{li2026rethinking} shows that successful OPD requires compatible student--teacher thinking patterns, genuinely new teacher knowledge, and increasing overlap between their high-probability token sets. Other studies identify failures caused by noisy student prefixes, entropy mismatch, unstable targets, or weak alignment between the student's reasoning state and teacher feedback~\cite{fu2026revisiting,jang2026stable,jin2026entropy,kim2026does}. These findings suggest that OPD depends not only on the divergence, but also on the \emph{context} where teacher supervision is applied. This further relates to the token-level loss direction: classical KD emphasizes teacher-confident tokens~\cite{hinton2015kd}, whereas many OPD implementations supervise tokens sampled from, or highly weighted by, the student policy~\cite{lu2025opd,xiao2026mimo}. 

\paragraph{Rollout selection, privileged hints, and data filtering.}
Our method is related to best-of-$N$ sampling, rejection sampling, and self-training methods that generate multiple candidates and retain high-quality outputs for inference or training~\cite{cobbe2021gsm8k,stiennon2020summarize,nakano2021webgpt,zelikman2022star,dong2023raft,lightman2023lets,wang2023selfconsistency,singh2024beyond}. Unlike these outer-loop filtering approaches, BRTS performs selection inside the OPD loop: from a small set of teacher rollouts, it chooses an auxiliary trajectory by prioritizing correctness and then proximity to the current student. The selected rollout is thus used not as offline training data, but as a teacher-context signal tailored to the current OPD update.
BRTS also connects to methods that use privileged information, such as ground-truth answers, demonstrations, or feedback, to strengthen supervision~\cite{snell2022learning,shenfeld2026sdft,hubotter2026rlvsd,zhao2026sdr,penaloza2026privileged,ding2026hdpo,yang2026selfrlvr}. 

\begin{figure*}[t]
    \centering
    \includegraphics[width=\textwidth]{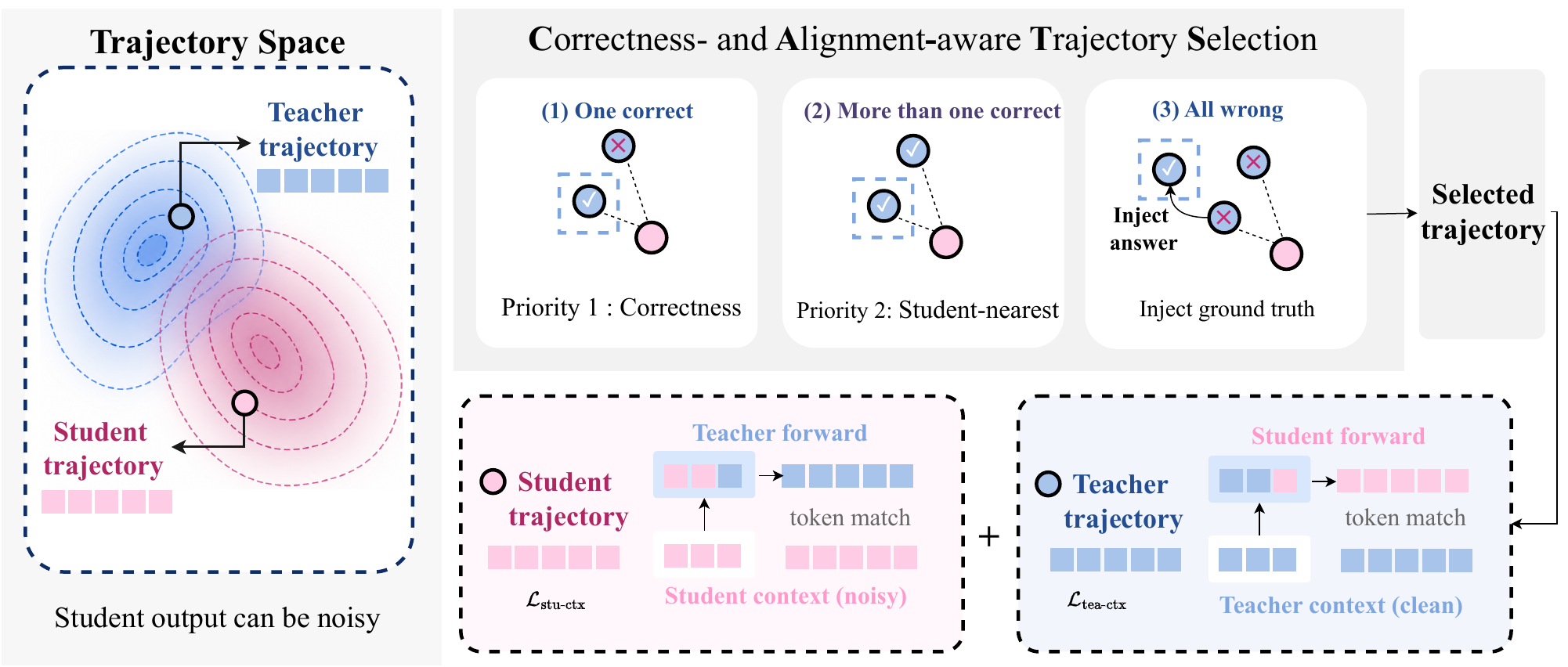}
    \caption{
    Overview of BRTS. The left panel shows teacher and student trajectories in the trajectory space. The upper-right panel illustrates the selection rule: choose a correct teacher rollout when available, select the student-nearest one when multiple correct rollouts exist, and inject the ground-truth answer when all sampled rollouts are wrong. The lower panels describe the loss computation, where token matching is performed under noisy student context and clean teacher context, respectively.
    }
    \label{fig:pipeline}
\end{figure*}

\section{Method}
\label{sec:method}

\subsection{Preliminaries}
\label{sec:prelim}
Let $x$ be an input prompt and $y^\star$ be its corresponding ground-truth answer. We consider a student policy $\pi_S$ and a teacher policy $\pi_T$, each defining an autoregressive next-token distribution over a vocabulary $\mathcal{V}$.
For a trajectory $y=(y_1,\ldots,y_T)$, the probability under a policy $\pi$ factorizes as
\begin{equation}
    \pi(y \mid x) = \prod_{t=1}^{T} \pi(y_t \mid x, y_{<t}),
\end{equation}
where $y_{<t}$ denotes the prefix before token $t$. Let $\hat{y}^S \sim \pi_S(\cdot \mid x)$ and $y^T \sim \pi_T(\cdot \mid x)$ denote student and teacher rollouts, respectively. On-policy distillation trains the student on states induced by its own generation: for rollout $\hat{y}^S$, the teacher provides token-level supervision on prefixes $(x,\hat{y}^S_{<t})$. A standard formulation minimizes the sequence-level reverse-KL objective~\cite{agarwal2024gkd,gu2023minillm,li2026rethinking}:

\begin{equation}
\label{eq:opd}
\mathcal{L}_{\mathrm{OPD}}(S)
=
\mathbb{E}_{x,\,\hat{y}^S\sim\pi_S}
\left[
\sum_{t=1}^{T}
D_{\mathrm{KL}}\!\left(
\pi_S(\cdot \mid x,\hat{y}^S_{<t})
\,\middle\|\,
\pi_T(\cdot \mid x,\hat{y}^S_{<t})
\right)
\right].
\end{equation}
In OPD implementations, the per-step KL is often approximated with a sampled or top-$K$ token set, typically chosen from the student distribution at the current student prefix~\cite{lu2025opd,xiao2026mimo}. 
This corrects the student on the states it actually visits.
However, the student context can be noisy.
BRTS addresses this by adding a correctness- and alignment-aware teacher-context branch.
%
%
We describe BRTS in three parts: teacher trajectory curation (Sec.~\ref{sec:BRTS}), teacher-context supervision (Sec.~\ref{sec:aux}), and the top-$K$ candidate direction used in each branch (Sec.~\ref{sec:topk}). 
Figure~\ref{fig:pipeline} shows our framework and Algorithm~\ref{alg:BRTS} summarizes the training step.

\subsection{Correctness- and Alignment-aware Trajectory Curation}
\label{sec:BRTS} 
In the first tier, we draw $N$ unconditioned teacher samples
$\{y^{T,i}\}_{i=1}^N \sim \pi_T(\cdot \mid x)$ for each prompt.
Each rollout is evaluated against the ground-truth answer $y^\star$ by extracting its final answer.
If at least one rollout is correct, \textit{i.e.}, $\mathrm{answer}(y^{T,i})=y^\star$, BRTS selects among the correct rollouts the trajectory with the highest token-level overlap with the student's top-$K$ candidate set.
This alignment criterion favors correct trajectories that remain close to the student's high-probability region and are therefore more suitable for distillation.
If none of the unconditioned teacher rollouts is correct, BRTS invokes a ground-truth-guided recovery step.
We construct a modified prompt $x^{\mathrm{gt}}$ that provides the correct answer $y^\star$ as a silent validation signal while instructing the teacher to produce a natural derivation.
A guided teacher rollout $y^{T,\mathrm{gt}} \sim \pi_T(\cdot \mid x^{\mathrm{gt}})$ is then sampled and retained only if its extracted answer is correct.
This step targets hard prompts where ordinary teacher sampling fails, allowing the teacher-context branch to remain active on examples where reliable supervision is most needed.
If neither unconditioned sampling nor guided recovery yields a correct trajectory, BRTS falls back to the Tier-1 rollout with the highest student overlap.
Although this fallback trajectory may be incorrect, selecting the closest available rollout avoids introducing an arbitrary teacher trajectory far from the student's distribution.

\begin{algorithm}[!t]
\caption{BRTS training step}
\label{alg:BRTS}
\begin{algorithmic}[1]
\REQUIRE Prompt $x$, ground-truth answer $y^\star$, teacher policy $\pi_T$, student policy $\pi_S$, number of Tier-1 rollouts $N$, auxiliary weight $\lambda$
\STATE Sample a student rollout $\hat{y}^S \sim \pi_S(\cdot \mid x)$
\STATE Draw $N$ unconditioned teacher rollouts $\{y^{T,i}\}_{i=1}^{N} \sim \pi_T(\cdot \mid x)$
\STATE Grade each teacher rollout by checking whether $\mathrm{answer}(y^{T,i}) = y^\star$
\IF{at least one Tier-1 rollout is correct}
    \STATE 
   Select $y'$ as the correct rollout with the highest top-$K$ overlap with the student trajectory.
\ELSE
    \STATE Construct the ground-truth-guided prompt $x^{\mathrm{gt}}$
    \STATE Sample one guided teacher rollout $y^{T,\mathrm{gt}} \sim \pi_T(\cdot \mid x^{\mathrm{gt}})$
    \IF{$\mathrm{answer}(y^{T,\mathrm{gt}})=y^\star$}
        \STATE Select $y' \gets y^{T,\mathrm{gt}}$
    \ELSE
        \STATE Select $y'$ as the Tier-1 rollout with the highest student top-$K$ overlap
    \ENDIF
\ENDIF
\STATE Compute the student-context loss $\mathcal{L}_{\mathrm{stu\text{-}ctx}}$ on $\hat{y}^S$ using student top-$K$ candidates
\STATE Compute the teacher-context loss $\mathcal{L}_{\mathrm{tea\text{-}ctx}}$ on the selected trajectory $y'$ using teacher top-$K$ candidates
\STATE $\mathcal{L}_{\mathrm{total}} \gets \mathcal{L}_{\mathrm{stu\text{-}ctx}} + \lambda \mathcal{L}_{\mathrm{tea\text{-}ctx}}$
\STATE Update the student parameters by taking a gradient step on $\mathcal{L}_{\mathrm{total}}$
\end{algorithmic}
\end{algorithm}

\subsection{Teacher-Context Supervision}
\label{sec:aux}

We retain Eq.~\eqref{eq:opd} on student-generated trajectories and augment it with a teacher-context distillation loss defined on the curated teacher trajectory $y'$ from Sec.~\ref{sec:BRTS}.
Both branches compare teacher and student distributions under matched conditioning contexts.
In the student-context branch, both distributions are conditioned on the student prefix $\hat{y}^S_{<t}$.
In the teacher-context branch, both distributions are conditioned on the selected teacher prefix $y'_{<t}$.
The resulting objectives are
\begin{align}
\mathcal{L}_{\text{stu-ctx}} &\;=\; \mathbb{E}\!\left[\sum_{t} D_{\mathrm{KL}}\!\big(\pi_S(\cdot \mid x, \hat{y}^S_{<t}) \,\|\, \pi_T(\cdot \mid x, \hat{y}^S_{<t})\big)\right], \label{eq:stu_ctx}\\
\mathcal{L}_{\text{tea-ctx}} &\;=\; \mathbb{E}\!\left[\sum_{t} D_{\mathrm{KL}}\!\big(\pi_T(\cdot \mid x, y'_{<t}) \,\|\, \pi_S(\cdot \mid x, y'_{<t})\big)\right], \label{eq:tea_ctx}\\
\mathcal{L}_{\text{total}} &\;=\; \mathcal{L}_{\text{stu-ctx}} + \lambda \, \mathcal{L}_{\text{tea-ctx}}. \label{eq:total}
\end{align}

The two branches play complementary roles.
The student-context loss preserves the on-policy nature of OPD by correcting the states visited by the student.
The teacher-context loss exposes the student to a coherent teacher reasoning path selected for both correctness and compatibility with the student's current distribution.
Thus, BRTS does not replace OPD; it augments OPD with a structured teacher-context signal that is less sensitive to stochastic teacher failures.
The coefficient $\lambda$ controls the strength of the teacher-context branch.
Since this branch is evaluated on selected teacher trajectories that are usually more coherent than noisy student rollouts, its raw contribution can be small with the naive choice $\lambda=1$.
Empirically, we find that $\lambda=10$ gives the auxiliary branch a meaningful scale while remaining stable, and use this value across all experiments.

\subsection{Top-$K$ Direction: Student- vs.\ Teacher-Confident}
\label{sec:topk}
In the sampled-token setting, both objectives are implemented as top-$K$ aggregation losses, but they differ in how the candidate token set is defined. For the student-context branch, the candidate set consists of the student's top-$K$ tokens under the student prefix. This branch therefore supervises the tokens that the student considers plausible, providing targeted correction on student-visited states.
For the teacher-context branch, the candidate set consists of the teacher's top-$K$ tokens under the selected teacher prefix. Unlike the student-context branch, this set reflects the teacher's local distribution along a high-quality trajectory. Thus, the auxiliary branch introduces teacher-preferred tokens that may lie outside the student's current top choices, while still keeping the supervision concentrated on a compact top-$K$ candidate set.

\begin{table*}[t]
\centering
\caption{\textbf{Effect of the teacher-context supervision.}
The baseline uses two student rollouts. Our variants replace one student rollout with one auxiliary teacher trajectory selected from $N$ teacher candidates. We report mean, best, and majority accuracy on AIME24, AIME25, and AMC23.}
\label{tab:main_results}
\small
\resizebox{\textwidth}{!}{
\begin{tabular}{lcc|ccc|ccc|ccc}
\toprule
\multicolumn{3}{c|}{Rollout Setup}
& \multicolumn{3}{c|}{AIME24}
& \multicolumn{3}{c|}{AIME25}
& \multicolumn{3}{c}{AMC23} \\
\cmidrule(lr){1-3}
\cmidrule(lr){4-6}
\cmidrule(lr){7-9}
\cmidrule(lr){10-12}
$\mathcal{L}_{\text{tea-ctx}}$
& Student
& Teacher / Cand.
& Mean & Best & Majority
& Mean & Best & Majority
& Mean & Best & Majority \\
\midrule

$\times$ 
& 2 &0 / 0
& 0.3917 & 0.5915 & 0.4146
& 0.2667 & 0.3503 & 0.2792
& 0.6777 & 0.7959 & 0.6952 \\

$\checkmark$ 
& 1 & 1 / 1
& 0.3750 & 0.5805 & 0.4124
& 0.2583 & 0.3823 & 0.2799
& 0.6747 & 0.8047 & \textbf{0.6978} \\

$\checkmark$ 
& 1 &1 / 2 
& 0.3750 & 0.5855 & 0.3953
& 0.2667 & \textbf{0.4056} & 0.2748
& 0.6717 & \textbf{0.8104} & 0.6917 \\

$\checkmark$ 
& 1 &1 / 4 
& \textbf{0.4000} & \textbf{0.5991} & \textbf{0.4306}
& \textbf{0.2917} & 0.3700 & \textbf{0.2961}
& \textbf{0.6837} & 0.8004 & 0.6948 \\


\bottomrule
\end{tabular}}
\end{table*}

\definecolor{studentColor}{RGB}{255,204,230}      
\definecolor{teacherColor}{RGB}{255,181,112}       
\begin{figure}[t]
    \centering
    \pgfplotsset{
        every axis/.style={
            width=\linewidth,
            height=6cm,
            xlabel={Training step},
            grid=both,
            major grid style={dashed, gray!30},
            minor grid style={dotted, gray!15},
            tick label style={font=\footnotesize},
            label style={font=\footnotesize},
            title style={font=\small, yshift=-0.5ex},
            axis line style={gray!60},
            tick style={gray!60},
            yticklabel style={/pgf/number format/.cd, fixed, precision=2, zerofill},
        },
    }
    \begin{subfigure}[t]{0.49\linewidth}
        \centering
        \begin{tikzpicture}
            \begin{axis}[
                ylabel={majority},
                xmin=5, xmax=55,
                xtick={10,20,30,40,50},
                ymin=0.37, ymax=0.44,
                ytick={0.37,0.38,0.39,0.40,0.41,0.42,0.43,0.44},
                legend pos=south east,
                legend style={
                    font=\footnotesize,
                    draw=none,
                    fill=white,
                    fill opacity=0.85,
                    text opacity=1,
                },
                legend cell align={left},
            ]
            \addplot[color=studentColor, mark=*, mark size=2pt, thick, mark options={fill=studentColor, draw=studentColor}] coordinates {
                (10,0.3868) (20,0.3866) (30,0.3957) (40,0.4234) (50,0.4166)
            };
            \addlegendentry{Baseline}
            \addplot[color=teacherColor, mark=triangle*, mark size=2.5pt, thick, mark options={fill=teacherColor, draw=teacherColor}] coordinates {
                (10,0.3953) (20,0.3932) (30,0.4277) (40,0.4277) (50,0.4225)
            };
            \addlegendentry{Teacher Cand. 2}
            \end{axis}
        \end{tikzpicture}
        \caption{Average over three benchmarks}
        \label{fig:avg_three}
    \end{subfigure}
    \hfill
    \begin{subfigure}[t]{0.49\linewidth}
        \centering
        \begin{tikzpicture}
            \begin{axis}[
                ylabel={majority},
                xmin=5, xmax=115,
                xtick={10,20,30,40,50,60,70,80,90,100,110},
                xticklabel style={font=\scriptsize},
                ymin=0.19, ymax=0.29,
                ytick={0.20,0.22,0.24,0.26,0.28},
                legend pos=south east,
                legend style={
                    font=\footnotesize,
                    draw=none,
                    fill=white,
                    fill opacity=0.85,
                    text opacity=1,
                },
                legend cell align={left},
            ]
            \addplot[color=studentColor, mark=*, mark size=2pt, thick, mark options={fill=studentColor, draw=studentColor}] coordinates {
                (10,0.2333) (20,0.2083) (30,0.2000) (40,0.2250) (50,0.2667)
                (60,0.2500) (70,0.2333) (80,0.2500) (90,0.2417) (100,0.2083)
                (110,0.2250)
            };
            \addlegendentry{Baseline}
            \addplot[color=teacherColor, mark=triangle*, mark size=2.5pt, thick, mark options={fill=teacherColor, draw=teacherColor}] coordinates {
                (10,0.2167) (20,0.2083) (30,0.2333) (40,0.2250) (50,0.2333)
                (60,0.2750) (70,0.2167) (80,0.2500) (90,0.2583) (100,0.2250)
                (110,0.2750)
            };
            \addlegendentry{Teacher Cand. 4}
            \end{axis}
        \end{tikzpicture}
        \caption{AIME 24 (teacher peaks higher)}
        \label{fig:aime_only}
    \end{subfigure}
    \caption{Majority-vote accuracy across training steps.
The student-only baseline uses two student rollouts, while BRTS replaces one student rollout with an auxiliary teacher trajectory selected from a small candidate pool. Left: average over AIME24, AIME25, and AMC23. Right: AIME24 results. BRTS achieves a higher early-training peak than the student-only baseline.}
    \label{fig:student_vs_teacher}
\end{figure}
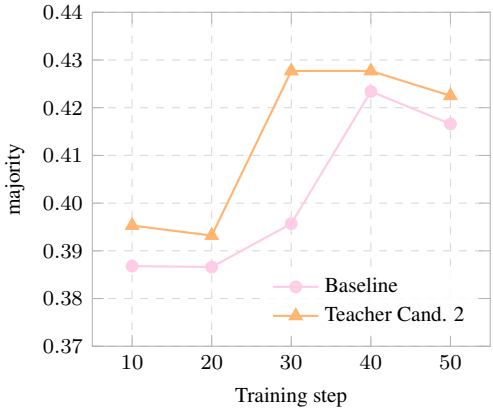
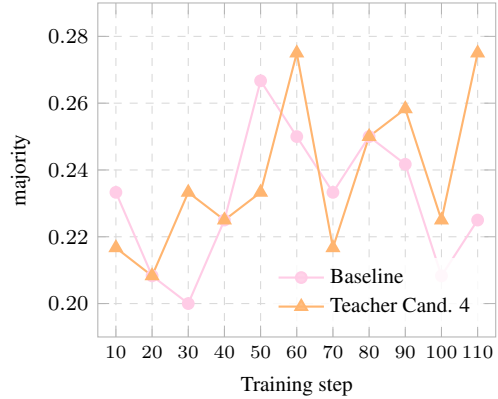

\section{Experiments}
\label{sec:experiments}
 
\subsection{Setup}
 
We follow the standard OPD training recipe used in recent work~\cite{lu2025opd,li2026rethinking}. Our main experiments use JustRL-1.5B~\cite{he2025justrl} as the teacher and a same-scale student backbone DeepSeek-1.5B~\cite{guo2025DeepSeekr1}, trained on instructions sampled from DAPO-Math-17K~\cite{yu2025dapo}. To examine whether the selector depends on a specific teacher family, we further conduct a teacher-swap experiment with DeepSeek-R1-Distill-Qwen-7B~\cite{guo2025DeepSeekr1} as the teacher. 
We evaluate mathematical reasoning on AIME 2024~\cite{li2024numinamath}, AIME 2025~\cite{balunovic2025matharena}, and AMC 2023~\cite{li2024numinamath}. AIME provides challenging short-answer problems requiring multi-step reasoning and exact numerical answers, while AMC 2023 offers relatively easier multiple-choice problems covering broader contest-math topics. We include both AIME 2024 and 2025 to test robustness across contest years, with AIME 2025 serving as a more difficult evaluation set.
By default, we sample $k=4$ solutions per problem with temperature $0.7$ and top-$p=0.95$, and report the average accuracy across $k$ sampled solutions, the best-of-$k$ accuracy, and the majority-vote accuracy.
More details are provided in Appendix~\ref{app:training}.


\subsection{Experiment Results}
\label{sec:main}

\begin{table*}[t]
\centering
\caption{\textbf{Effect of Tier-2 ground-truth-guided selection.}
Tier 1 uses unconditioned teacher samples, while Tier 2 adds ground-truth-guided re-sampling for harder prompts not solved by Tier 1. We report mean, best, and majority-vote accuracy on AIME25 and AMC23.}
\label{tab:tier2}
\small
\resizebox{\textwidth}{!}{
\begin{tabular}{cccccc|ccc|ccc}
\toprule
\multirow{2}{*}{$\mathcal{L}_{\text{tea-ctx}}$} & \multirow{2}{*}{Tier 2}&\multicolumn{4}{c|}{Rollout Setup}
& \multicolumn{3}{c|}{AIME25}
& \multicolumn{3}{c}{AMC23} \\
\cmidrule(lr){3-6}
\cmidrule(lr){7-9}
\cmidrule(lr){10-12}
& & Student
& Teacher / Cand. & Tier 1 & Tier 2
& Mean & Best & Majority
& Mean & Best & Majority \\
\midrule

$\times$ 
& $\times$ & 2 &0 / 0 & 0 & 0
& 0.2667 & 0.3503 & 0.2792
& 0.6777 & 0.7959 & 0.6952 \\

$\checkmark$ 
&$\times$ & 1 &1 / 2 & 2 & 0
& 0.2667 & 0.4056 & 0.2748
& 0.6717 & \textbf{0.8104} & 0.6917 \\

$\checkmark$ 
& $\times$ & 1 &1 / 4 & 4 & 0
& 0.2917 & 0.3700 & 0.2961
& \textbf{0.6837} & 0.8004 & 0.6948 \\

$\checkmark$ & $\checkmark$  & 1 & 1 / 2+1 & 2 & 1
& \textbf{0.3000} & \textbf{0.4309} & \textbf{0.3133}
& 0.6747 & 0.8066 & \textbf{0.6997} \\

\bottomrule
\end{tabular}}
\end{table*}

\begin{figure}[t]
    \centering

    \definecolor{mayablue}{RGB}{255,181,112}
    \definecolor{geraldine}{RGB}{94,169,248}
    \definecolor{gridcol}{RGB}{210,208,200}
    \definecolor{axiscol}{RGB}{180,178,169}
    \definecolor{labelcol}{RGB}{95,94,90}

    \pgfplotsset{
        every axis/.style={
            width=\linewidth,
            height=6.2cm,
            xlabel={Training step},
            xmin=0, xmax=285,
            xtick={0,40,80,120,160,200,240,280},
            grid=both,
            major grid style={thin, gridcol},
            minor grid style={thin, gridcol!50},
            tick label style={font=\footnotesize, color=labelcol},
            label style={font=\footnotesize, color=labelcol},
            title style={font=\small\bfseries, color=black, yshift=2pt},
            axis line style={axiscol, line width=0.6pt},
            tick style={axiscol, line width=0.6pt},
            yticklabel style={
                color=labelcol,
                /pgf/number format/.cd,
                fixed, precision=2, zerofill,
            },
            enlarge x limits=0.02,
            clip=true,
            legend style={
                font=\footnotesize,
                draw=none,
                fill=white,
                fill opacity=0.85,
                text opacity=1,
                row sep=1pt,
                inner sep=3pt,
                at={(0.98,0.98)},
                anchor=north east,
            },
            legend cell align={left},
        },
    }

    \begin{subfigure}[t]{0.49\linewidth}
        \centering
        \begin{tikzpicture}
            \begin{axis}[
                ymin=0.20, ymax=0.32,
                ytick={0.20,0.22,0.24,0.26,0.28,0.30,0.32},
            ]

            \addplot[
                color=geraldine,
                mark=square*,
                mark size=1.8pt,
                thick,
                mark options={fill=geraldine, draw=geraldine},
            ] coordinates {
                (10,0.2718) (20,0.3117) (30,0.2594) (40,0.2207) (50,0.2492)
                (60,0.2550) (70,0.2182) (80,0.2336) (90,0.2547) (100,0.3069)
                (110,0.2585) (120,0.2572) (130,0.2257) (140,0.2632)
                (160,0.2310) (170,0.2253) (180,0.3133) (190,0.2696) (200,0.2478)
                (210,0.2382) (220,0.2818) (230,0.2729) (240,0.2638) (250,0.2517)
                (260,0.2062) (270,0.2400) (279,0.2287)
            };
            \addlegendentry{w/ tier-2}

            \addplot[
                color=mayablue,
                mark=triangle*,
                mark size=2.5pt,
                thick,
                mark options={fill=mayablue, draw=mayablue, solid},
            ] coordinates {
                (10,0.2323) (20,0.2309) (30,0.2398) (40,0.2438) (50,0.2404)
                (60,0.2609) (70,0.2350) (80,0.2130) (90,0.2384) (100,0.2461)
                (110,0.2560) (120,0.2711) (130,0.2565) (140,0.2234) (150,0.2534)
                (160,0.2521) (170,0.2650) (180,0.2748) (190,0.2620) (200,0.2740)
                (210,0.2682) (220,0.2690) (230,0.2224) (240,0.2550) (250,0.2392)
                (260,0.2590) (270,0.2329) (279,0.2073)
            };
            \addlegendentry{w/o tier2}

            \end{axis}
        \end{tikzpicture}
        \caption{AIME 25}
        \label{fig:aime25}
    \end{subfigure}
    \hfill
    \begin{subfigure}[t]{0.49\linewidth}
        \centering
        \begin{tikzpicture}
            \begin{axis}[
                ymin=0.24, ymax=0.42,
                ytick={0.25,0.30,0.35,0.40},
            ]

            \addplot[
                color=geraldine,
                mark=square*,
                mark size=1.8pt,
                thick,
                mark options={fill=geraldine, draw=geraldine},
            ] coordinates {
                (10,0.3196) (20,0.2824) (30,0.3861) (40,0.3672) (50,0.3742)
                (60,0.2951) (70,0.3524) (80,0.2887) (90,0.2920) (100,0.3953)
                (110,0.3237) (120,0.3136) (130,0.3472) (140,0.2936) (150,0.3566)
                (160,0.3210) (170,0.3021) (180,0.3282) (190,0.3815) (200,0.3714)
                (210,0.3275) (220,0.3595) (230,0.3286) (240,0.3761) (250,0.3174)
                (260,0.3461) (270,0.3788) (279,0.2625)
            };
            \addlegendentry{w/ tier-2}

            \addplot[
                color=mayablue,
                mark=triangle*,
                mark size=2.5pt,
                thick,
                mark options={fill=mayablue, draw=mayablue, solid},
            ] coordinates {
                (10,0.2964) (20,0.2739) (30,0.3123) (40,0.3382) (50,0.2877)
                (60,0.3342) (70,0.3105) (80,0.3204) (90,0.3175) (100,0.2630)
                (110,0.3198) (120,0.2949) (130,0.3152) (140,0.3358) (150,0.3426)
                (160,0.2965) (170,0.2913) (180,0.2699) (190,0.3397) (200,0.3651)
                (210,0.3723) (220,0.3486) (230,0.3415) (240,0.3426) (250,0.3379)
                (260,0.3327) (270,0.3544) (279,0.2981)
            };
            \addlegendentry{w/o tier2}

            \end{axis}
        \end{tikzpicture}
        \caption{AIME 24}
        \label{fig:aime24}
    \end{subfigure}

    \caption{%
        Majority accuracy on AIME25 and AIME24 across training steps, comparing Tier-1 selection with two teacher candidates to the same setting augmented with Tier-2 recovery. Tier-2 substantially improves performance on the harder AIME25 benchmark, suggesting that teacher-guided correction is most beneficial for difficult tasks.
    }
    \label{fig:aime_maj4}
\end{figure}
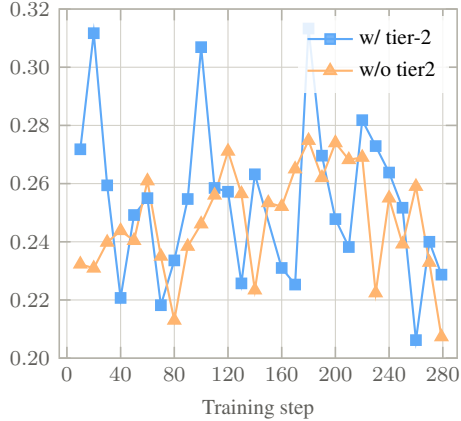
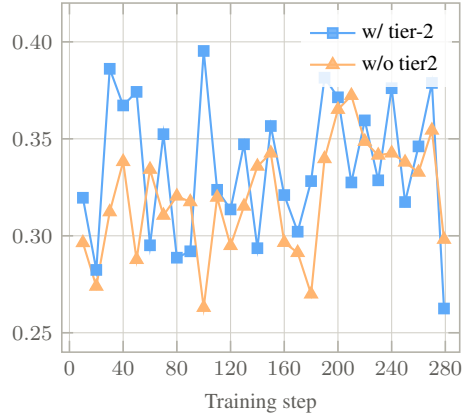
 
 
 

\paragraph{Effect of the teacher-context loss.}
We first evaluate whether teacher rollouts provide additional supervision beyond the standard student-only rollout. Table~\ref{tab:main_results} compares different rollout configurations on AIME24, AIME25, and AMC23.
Overall, teacher rollouts improve performance on the more challenging AIME benchmarks. Compared with the student-only setting, replacing one student rollout with a selected teacher rollout and enlarging the teacher candidate pool generally yields more informative supervision. For example, using four teacher candidates achieves the best AIME24 results, reaching $0.400$ average accuracy, $0.599$ best accuracy, and $0.4306$ majority-vote accuracy. This suggests that sampling multiple teacher trajectories increases the likelihood of selecting a useful reasoning path.
Figure~\ref{fig:student_vs_teacher} further shows that replacing one student rollout with a teacher rollout leads to a substantially higher early-stage performance peak. This indicates that a selected teacher rollout can provide more informative supervision than an additional student rollout.
On AMC23, the gains are more modest, likely because the student-only baseline is already strong. Nevertheless, using four teacher rollout candidates still improves average accuracy, suggesting that a larger teacher candidate pool can provide useful supervision even on easier benchmarks.

\paragraph{Effect of Tier-2 ground-truth-guided selection.} 
We further analyze Tier-2 selection, where a ground-truth-guided hint is injected into the teacher prompt when all initial teacher samples fail. Figures~\ref{fig:aime25} and~\ref{fig:aime24} compare training with and without Tier-2 on AIME25 and AIME24. On the harder AIME25 benchmark, Tier-2 consistently improves majority-vote accuracy at early training stages. It also achieves the best results in Table~\ref{tab:tier2}, reaching $0.3000$ average accuracy, $0.4309$ best accuracy, and $0.3133$ majority-vote accuracy. This suggests that hint-guided teacher correction is particularly useful when unconditioned teacher rollouts fail to provide reliable supervision.

\paragraph{Computational Cost.}
Increasing the teacher candidate pool mainly affects the selection stage: each additional candidate adds about $59$ seconds per step in our setup, as detailed in Appendix~\ref{app:additional}, which remains acceptable. Importantly, it does not increase the number of trajectories used in the loss. In the two-rollout setting, for example, BRTS replaces one student rollout with one selected teacher rollout. Since teacher solutions are often more confident and direct, they tend to produce shorter sequences.
BRTS also improves the compute-performance trade-off by achieving stronger performance in early training. As shown in Figures~\ref{fig:student_vs_teacher} and~\ref{fig:aime_maj4}, replacing one student rollout with a selected teacher rollout leads to a higher early-training peak, and Tier-2 further strengthens this effect on harder tasks. Since longer distillation may suffer from catastrophic forgetting, reaching better performance earlier can enable shorter training and reduce total computation. Thus, BRTS trades a modest teacher-sampling overhead for higher-quality supervision, faster convergence, and a higher potential performance upper bound.

\begin{table}[!t]
\centering
\caption{Effect of applying a prompt perturbation during teacher rollout, on AMC 2023.}
\label{tab:disturb}
\small
\begin{tabular}{llccccc}
\toprule
\textbf{Metric} & \textbf{Method} & \textbf{Step 10} & \textbf{Step 20} & \textbf{Step 30} & \textbf{Step 40} & \textbf{Step 50} \\
\midrule
\multirow{2}{*}{AMC mean}
  & w/o perturb & 0.6175 & 0.6416 & 0.6416 & 0.6446 & 0.6325 \\
  & w/ perturb  & \textbf{0.6566} & \textbf{0.6566} & 0.6476 & 0.6265 & 0.6476 \\
\midrule
\multirow{2}{*}{AMC majority}
  & w/o perturb & 0.6339 & 0.6663 & 0.6570 & 0.6721 & 0.6531 \\
  & w/ perturb  & 0.6759 & \textbf{0.6764} & 0.6615 & 0.6360 & 0.6698 \\
\bottomrule
\end{tabular}
\end{table}

\pgfplotsset{compat=1.18}
\usepgfplotslibrary{fillbetween}
\usetikzlibrary{patterns, calc}

\definecolor{observedred}{RGB}{192, 57, 43}
\definecolor{baselinegray}{RGB}{130, 130, 130}
\definecolor{disturbgreen}{RGB}{76, 153, 134}

\pgfplotsset{
    panel/.style={
        width=8cm, height=6cm,
        xlabel={Total rollouts per prompt},
        ylabel={Tier-1 accuracy},
        xmin=0.7, xmax=4.5,
        xtick={1,2,3,4},
        grid=major,
        grid style={gray!20, line width=0.3pt},
        axis line style={gray!50},
        tick style={gray!50},
        label style={font=\small},
        tick label style={font=\footnotesize},
        legend cell align={left},
    },
}

\begin{figure}[t]
    \centering
    \resizebox{\textwidth}{!}{%
    \begin{tikzpicture}
    \begin{axis}[
        name=left,
        panel,
        ymin=0, ymax=1.0,
        ytick={0,0.2,0.4,0.6,0.8,1.0},
        yticklabels={0\%,20\%,40\%,60\%,80\%,100\%},
        legend style={
            font=\scriptsize,
            at={(0.5,1.03)}, anchor=south,
            legend columns=1,
            draw=none,
            row sep=1pt,
        },
    ]

    \addplot[
        color=baselinegray, line width=1.2pt, dashed,
        mark=o, mark size=3pt,
        mark options={solid, line width=1pt, fill=white},
    ] coordinates {
        (1, 0.4336) (2, 0.67919104)
        (3, 0.8182938051) (4, 0.8970816112)
    };
    \addlegendentry{i.i.d.\ baseline: $1-(1-p)^n$}

    \addplot[
        color=observedred, line width=1.5pt,
        mark=*, mark size=3pt, mark options={solid},
    ] coordinates {
        (1, 0.4336) (2, 0.5273)
        (3, 0.5859375) (4, 0.6055)
    };
    \addlegendentry{Observed accuracy}

    \node[font=\scriptsize, gray!80!black, anchor=north] at (axis cs:2, 0.5273) {\shortstack{$n{=}2$\\52.7\%}};
    \node[font=\scriptsize, gray!80!black, anchor=north] at (axis cs:3, 0.5859375) {\shortstack{$n{=}3$\\58.6\%}};
    \node[font=\scriptsize, gray!80!black, anchor=west] at (axis cs:4.05, 0.6055) {\shortstack{$n{=}4$\\60.5\%}};

    \node[font=\scriptsize, gray!80!black, anchor=north west, align=left] at (axis cs:1.15, 0.36)
        {\textit{single rollout:}\\\textit{43.4\% correct}};
    \draw[gray!60, line width=0.4pt] (axis cs:1.15, 0.36) -- (axis cs:1.02, 0.42);

    \end{axis}

    \begin{axis}[
        name=right,
        at={($(left.east)+(2cm,0)$)}, anchor=west,
        panel,
        ymin=0.4, ymax=1.0,
        ytick={0.4,0.5,0.6,0.7,0.8,0.9,1.0},
        yticklabels={40\%,50\%,60\%,70\%,80\%,90\%,100\%},
        legend style={
            font=\scriptsize,
            at={(0.5,1.03)}, anchor=south,
            legend columns=2,
            draw=none,
            row sep=1pt,
            column sep=6pt,
            /tikz/every even column/.append style={column sep=6pt},
        },
    ]

    \addplot[name path=disturb_path, draw=none, forget plot, smooth, tension=0.7] coordinates {
        (1, 0.4336) (2, 0.578125) (4, 0.65625)
    };
    \addplot[name path=observed_path, draw=none, forget plot, smooth, tension=0.7] coordinates {
        (1, 0.4336) (2, 0.5273) (3, 0.5859375) (4, 0.6055)
    };
    \addplot[disturbgreen!25, forget plot] fill between[of=disturb_path and observed_path];

    \addplot[
        color=baselinegray, line width=1.2pt, dashed,
        mark=square, mark size=2.5pt,
        mark options={solid, line width=1pt, fill=baselinegray!60},
        smooth, tension=0.7,
    ] coordinates {
        (1, 0.4336) (2, 0.67919104)
        (3, 0.8182938051) (4, 0.8970816112)
    };
    \addlegendentry{Theoretical: $1-(1-p)^n$}

    \addplot[
        color=disturbgreen, line width=1.5pt,
        mark=o, mark size=3pt,
        mark options={solid, line width=1.2pt, fill=white},
        smooth, tension=0.7,
    ] coordinates {
        (1, 0.4336) (2, 0.578125) (4, 0.65625)
    };
    \addlegendentry{Disturb}

    \addplot[
        color=observedred, line width=1.5pt, dashed,
        mark=triangle*, mark size=3pt, mark options={solid},
        smooth, tension=0.7,
    ] coordinates {
        (1, 0.4336) (2, 0.5273)
        (3, 0.5859375) (4, 0.6055)
    };
    \addlegendentry{Observed}

    \addplot[fill=disturbgreen!25, draw=disturbgreen!50, area legend] coordinates {(0,0)};
    \addlegendentry{Difference}

    \node[font=\scriptsize, disturbgreen!80!black, anchor=west] at (axis cs:2.55, 0.555) {\textit{difference}};

    \end{axis}

    \node[font=\small, anchor=north, yshift=-1.1cm] at (left.south) {(a) Observed vs.\ i.i.d.\ baseline.};
    \node[font=\small, anchor=north, yshift=-1.1cm] at (right.south) {(b) Disturbance vs.\ observed vs.\ theoretical.};

    \end{tikzpicture}%
    }
    \caption{Tier-1 accuracy as a function of total rollouts per prompt. Panel~(a) compares observed accuracy against the i.i.d.\ baseline $1-(1-p)^n$. Panel~(b) shows the gap between rollouts with one disturbance, the observed accuracy, and the theoretical i.i.d.\ curve.}
    \label{fig:rollout-comparison}
\end{figure}
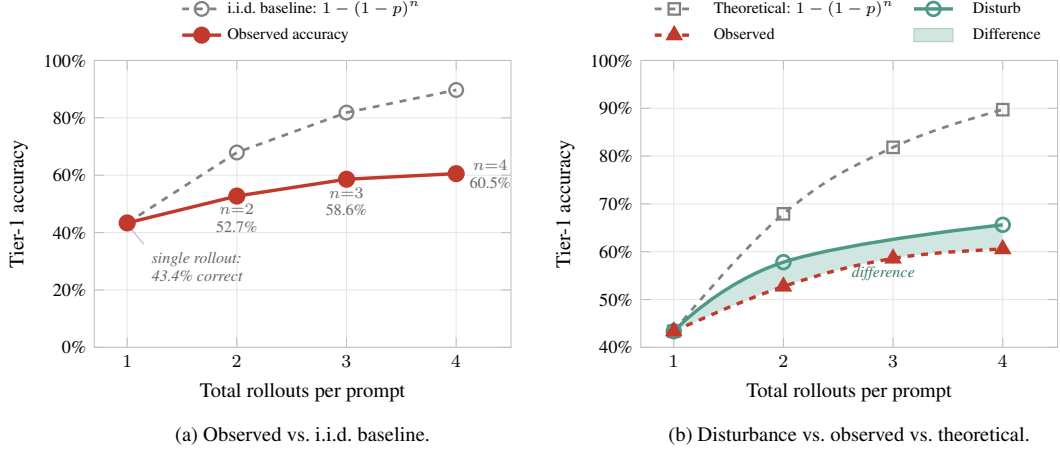

\paragraph{Prompt Perturbation.}
We further examine the sensitivity of BRTS to teacher prompt wording by applying a small prompt perturbation to one of the two teacher candidates. Detailed prompts are provided in Appendix~\ref{app:prompts}. As shown in Table~\ref{tab:disturb}, this perturbation does not degrade early performance on AMC23; instead, it raises early step best average accuracy from $0.6416$ to $0.6566$ and majority-vote accuracy from $0.6663$ to $0.6764$.
Figure~\ref{fig:rollout-comparison} helps explain this effect. Without added diversity, the Tier-1 catch rate grows much more slowly than the ideal i.i.d. prediction $1-(1-p)^n$, suggesting strong correlation among teacher samples. The perturbed rollout shifts the observed curve upward, indicating that mild prompt variation partially decorrelates teacher trajectories and increases the chance of finding a correct one.
Overall, BRTS is not brittle to a single teacher prompt. Since selection is correctness-filtered, mild prompt perturbation can safely improve rollout diversity while preserving reliable supervision, especially in early training.

\paragraph{Backbones.}
\label{sec:DeepSeek}
 
\begin{table}[t]
\centering
\caption{Backbone ablation with DeepSeek-R1-Distill-Qwen-7B as the teacher at early step.}
\label{tab:rm_reward_ablation_vs_changeopd}
\small
\resizebox{\textwidth}{!}{
\begin{tabular}{cc|ccc|ccc|ccc}
\toprule
\multicolumn{2}{c|}{Rollout Setup} & \multicolumn{3}{c|}{AIME24}
& \multicolumn{3}{c|}{AIME25}
& \multicolumn{3}{c}{AMC23}\\
\cmidrule(lr){1-2}
\cmidrule(lr){3-5}
\cmidrule(lr){6-8}
\cmidrule(lr){9-11}
Student & Teacher / Cand.
& Mean & Best & Maj
& Mean & Best & Maj
& Mean & Best & Maj
\\
\midrule
2 & 0 / 0
& 0.3167 &\textbf{0.4499} & 0.3331
& 0.2167 & 0.2770 & 0.2277
& 0.6265 & 0.7410 & 0.6453\\

1&1 / 2
& \textbf{0.3667} & 0.4485 & \textbf{0.3952}
& \textbf{0.2417} & \textbf{0.3294} & \textbf{0.2528}
& \textbf{0.6446} & \textbf{0.7836} & \textbf{0.6676}\\

\bottomrule
\end{tabular}}
\end{table}

\definecolor{n1color}{HTML}{F4A6A6}        
\definecolor{n1tier2color}{HTML}{B22222}   
\definecolor{n2color}{HTML}{E8B53D}        
\definecolor{n2tier2color}{HTML}{8C6B0F}   
\definecolor{n4color}{HTML}{6FAE7E}        
\definecolor{gridcolor}{HTML}{D8D8DC}
\definecolor{axiscolor}{HTML}{4A4A4A}

\definecolor{tier1blue}{RGB}{140,216,254}   
\definecolor{tier2yellow}{RGB}{254,228,111} 
\definecolor{tier2orange}{RGB}{254,228,111} 
\definecolor{tier3gray}{RGB}{179,179,179}

\def\barheight{0.50}
\def\barwidth{7.2}

\newcommand{\barrow}[5]{%
    \node[anchor=east, font=\sffamily\small] at (-0.20, #1) {#2};
    \pgfmathsetmacro{\tone}{#3/100*\barwidth}
    \fill[tier1blue] (0, #1-\barheight/2) rectangle (\tone, #1+\barheight/2);

    \pgfmathsetmacro{\ttwo}{#4/100*\barwidth}
    \pgfmathsetmacro{\tonetwo}{\tone+\ttwo}
    \ifdim #4pt>0pt
        \fill[tier2yellow] (\tone, #1-\barheight/2) rectangle (\tonetwo, #1+\barheight/2);
    \fi

    \pgfmathsetmacro{\tthree}{#5/100*\barwidth}
    \pgfmathsetmacro{\tend}{\tonetwo+\tthree}
    \fill[tier3gray] (\tonetwo, #1-\barheight/2) rectangle (\tend, #1+\barheight/2);

    \node[white, font=\sffamily\scriptsize\bfseries]
        at (\tone/2, #1) {#3\%};
    \ifdim #4pt>0pt
        \node[white, font=\sffamily\scriptsize\bfseries]
            at (\tone+\ttwo/2, #1) {#4\%};
    \fi
    \node[white, font=\sffamily\scriptsize\bfseries]
        at (\tonetwo+\tthree/2, #1) {#5\%};
}

\begin{figure*}[!t]
    \centering
    \begin{tikzpicture}
        \pgfplotsset{
            every axis/.style={
                width=0.38\textwidth,
                height=4.6cm,
                xmin=0.6, xmax=10.4,
                xtick={1,2,3,4,5,6,7,8,9,10},
                xticklabels={},        
                ymin=0.40, ymax=0.82,
                ytick={0.40,0.50,0.60,0.70,0.80},
                grid=major,
                major grid style={thin, gridcolor},
                tick label style={font=\scriptsize, /pgf/number format/fixed},
                label style={font=\footnotesize},
                title style={font=\small\normalfont, yshift=-3pt, align=center},
                axis line style={axiscolor, line width=0.5pt},
                tick style={axiscolor, line width=0.4pt},
                line width=1.4pt,
                mark options={solid, line width=0.7pt},
                yticklabel style={
                    /pgf/number format/.cd, fixed, precision=2, zerofill,
                },
                enlarge x limits=false,
            },
        }

        \begin{axis}[
            name=plotA,
            title={(a) Effect of T-1 candidates.},
            ylabel={Accuracy},
        ]
            \addplot[color=n1color, mark=*, mark size=1.7pt,
                     mark options={fill=n1color, draw=n1color!70!black}]
                coordinates {
                    (1,0.4336) (2,0.4609) (3,0.5313) (4,0.5469) (5,0.5313)
                    (6,0.5313) (7,0.4844) (8,0.5703) (9,0.5430) (10,0.5430)
                };
            \addplot[color=n2color, mark=square*, mark size=1.7pt,
                     mark options={fill=n2color, draw=n2color!70!black}]
                coordinates {
                    (1,0.5273) (2,0.5117) (3,0.6133) (4,0.6328) (5,0.6055)
                    (6,0.6289) (7,0.5586) (8,0.6445) (9,0.6016) (10,0.6445)
                };
            \addplot[color=n4color, mark=triangle*, mark size=2.2pt,
                     mark options={fill=n4color, draw=n4color!70!black}]
                coordinates {
                    (1,0.6055) (2,0.6250) (3,0.6875) (4,0.6836) (5,0.6680)
                    (6,0.6797) (7,0.6641) (8,0.7031) (9,0.6602) (10,0.6914)
                };
        \end{axis}

        \begin{axis}[
            name=plotB,
            at={($(plotA.east)+(0.6cm,0)$)}, anchor=west,
            title={(b) T-2 with one T-1 candidate.},
            yticklabels={}, ylabel={},
        ]
            \addplot[name path=baseA, draw=none, forget plot]
                coordinates {
                    (1,0.4336) (2,0.4609) (3,0.5313) (4,0.5469) (5,0.5313)
                    (6,0.5313) (7,0.4844) (8,0.5703) (9,0.5430) (10,0.5430)
                };
            \addplot[name path=tier2A, draw=none, forget plot]
                coordinates {
                    (1,0.6836) (2,0.6680) (3,0.7500) (4,0.6875) (5,0.7187)
                    (6,0.7344) (7,0.7109) (8,0.7305) (9,0.6602) (10,0.6875)
                };
            \addplot[n1tier2color!12, forget plot] fill between[of=baseA and tier2A];

            \addplot[color=n1color, mark=*, mark size=1.7pt,
                     mark options={fill=n1color, draw=n1color!70!black}]
                coordinates {
                    (1,0.4336) (2,0.4609) (3,0.5313) (4,0.5469) (5,0.5313)
                    (6,0.5313) (7,0.4844) (8,0.5703) (9,0.5430) (10,0.5430)
                };
            \addplot[color=n1tier2color, mark=diamond*, mark size=2.2pt,
                     mark options={fill=n1tier2color, draw=n1tier2color!70!black}]
                coordinates {
                    (1,0.6836) (2,0.6680) (3,0.7500) (4,0.6875) (5,0.7187)
                    (6,0.7344) (7,0.7109) (8,0.7305) (9,0.6602) (10,0.6875)
                };
        \end{axis}

        \begin{axis}[
            name=plotC,
            at={($(plotB.east)+(0.6cm,0)$)}, anchor=west,
            title={(c) T-2 with two T-1 candidates.},
            yticklabels={}, ylabel={},
            legend columns=5,
            legend to name=combinedlegend,
            legend style={
                font=\footnotesize,
                draw=none, fill=none,
                inner sep=2pt, column sep=10pt,
            },
            legend cell align={left},
        ]
            \addplot[name path=baseB, draw=none, forget plot]
                coordinates {
                    (1,0.5273) (2,0.5117) (3,0.6133) (4,0.6328) (5,0.6055)
                    (6,0.6289) (7,0.5586) (8,0.6445) (9,0.6016) (10,0.6445)
                };
            \addplot[name path=tier2B, draw=none, forget plot]
                coordinates {
                    (1,0.6953) (2,0.6641) (3,0.7852) (4,0.7187) (5,0.7305)
                    (6,0.7344) (7,0.7031) (8,0.7617) (9,0.6680) (10,0.6992)
                };
            \addplot[n2tier2color!12, forget plot] fill between[of=baseB and tier2B];

            \addplot[color=n1color, mark=*, mark size=1.7pt,
                     mark options={fill=n1color, draw=n1color!70!black}]
                coordinates { (1,0) };
            \addlegendentry{\# 1}

            \addplot[color=n2color, mark=square*, mark size=1.7pt,
                     mark options={fill=n2color, draw=n2color!70!black}]
                coordinates {
                    (1,0.5273) (2,0.5117) (3,0.6133) (4,0.6328) (5,0.6055)
                    (6,0.6289) (7,0.5586) (8,0.6445) (9,0.6016) (10,0.6445)
                };
            \addlegendentry{\# 2}

            \addplot[color=n4color, mark=triangle*, mark size=2.2pt,
                     mark options={fill=n4color, draw=n4color!70!black}]
                coordinates { (1,0) };
            \addlegendentry{\# 3}

            \addplot[color=n1tier2color, mark=diamond*, mark size=2.2pt,
                     mark options={fill=n1tier2color, draw=n1tier2color!70!black}]
                coordinates { (1,0) };
            \addlegendentry{\# 4}

            \addplot[color=n2tier2color, mark=diamond*, mark size=2.2pt,
                     mark options={fill=n2tier2color, draw=n2tier2color!70!black}]
                coordinates {
                    (1,0.6953) (2,0.6641) (3,0.7852) (4,0.7187) (5,0.7305)
                    (6,0.7344) (7,0.7031) (8,0.7617) (9,0.6680) (10,0.6992)
                };
            \addlegendentry{\# 5}
        \end{axis}

        \node[anchor=north] at ($(plotA.south west)!0.5!(plotC.south east)+(0,-0cm)$)
            {\pgfplotslegendfromname{combinedlegend}};

    \end{tikzpicture}

    \vspace{-0.4em}

\definecolor{tier1blue}{RGB}{92, 181, 238}
\definecolor{tier2yellow}{RGB}{247, 193, 48}
\definecolor{tier3gray}{RGB}{145, 150, 158}
\definecolor{axisgray}{RGB}{80, 80, 80}
\definecolor{gridgray}{RGB}{205, 205, 205}

\renewcommand{\barwidth}{6.8}
\renewcommand{\barheight}{0.52}
\newcommand{\barcorner}{2.8pt}

\newcommand{\prettybarrow}[5]{%
    \pgfmathsetmacro{\xone}{#3/100*\barwidth}
    \pgfmathsetmacro{\xtwo}{(#3+#4)/100*\barwidth}
    \pgfmathsetmacro{\xthree}{(#3+#4+#5)/100*\barwidth}

    \node[anchor=east, font=\sffamily\bfseries\Large, text=black!85]
        at (-0.32,#1) {#2};

    \begin{scope}
        \clip[rounded corners=\barcorner]
            (0,#1-\barheight/2) rectangle (\barwidth,#1+\barheight/2);

        \fill[tier1blue]
            (0,#1-\barheight/2) rectangle (\xone,#1+\barheight/2);

        \ifdim #4 pt > 0pt
            \fill[tier2yellow]
                (\xone,#1-\barheight/2) rectangle (\xtwo,#1+\barheight/2);
        \fi

        \fill[tier3gray]
            (\xtwo,#1-\barheight/2) rectangle (\barwidth,#1+\barheight/2);

        \fill[white, opacity=0.10]
            (0,#1+0.04) rectangle (\barwidth,#1+\barheight/2);
    \end{scope}

    \draw[white, line width=0.75pt]
        (\xone,#1-\barheight/2) -- (\xone,#1+\barheight/2);

    \ifdim #4 pt > 0pt
        \draw[white, line width=0.75pt]
            (\xtwo,#1-\barheight/2) -- (\xtwo,#1+\barheight/2);
    \fi

    \node[font=\sffamily\bfseries\large, text=white]
        at (0.5*\xone,#1) {#3\%};

    \ifdim #4 pt > 0pt
        \node[font=\sffamily\bfseries\large, text=white]
            at (0.5*(\xone+\xtwo),#1) {#4\%};
    \fi

    \node[font=\sffamily\bfseries\large, text=white]
        at (0.5*(\xtwo+\barwidth),#1) {#5\%};
}

\renewcommand{\barwidth}{6.8}
\renewcommand{\barheight}{0.50}
\newcommand{\barsegsep}{0.1}   
\newcommand{\barradius}{4pt}

\definecolor{tier1blue}{RGB}{76,114,176}   
\definecolor{tier2yellow}{RGB}{221,170,51} 
\definecolor{tier3gray}{RGB}{170,170,170}  
\renewcommand{\barrow}[5]{%
    \pgfmathsetmacro{\xA}{#3/100*\barwidth}
    \pgfmathsetmacro{\xB}{(#3+#4)/100*\barwidth}

    \node[
        anchor=east,
        font=\sffamily\bfseries\fontsize{13}{13}\selectfont
    ] at (-0.42,#1) {#2};

    \begin{scope}
        \clip[rounded corners=\barradius]
            (0,#1-\barheight/2) rectangle (\barwidth,#1+\barheight/2);

        \ifdim #4 pt > 0pt
            \fill[tier1blue]
                (0,#1-\barheight/2)
                rectangle ({\xA-\barsegsep/2},#1+\barheight/2);

            \fill[tier2yellow]
                ({\xA+\barsegsep/2},#1-\barheight/2)
                rectangle ({\xB-\barsegsep/2},#1+\barheight/2);

            \fill[tier3gray]
                ({\xB+\barsegsep/2},#1-\barheight/2)
                rectangle (\barwidth,#1+\barheight/2);

            \node[
                font=\sffamily\bfseries\fontsize{11}{11}\selectfont,
                text=white
            ] at ({(\xA-\barsegsep/2)/2},#1) {#3\%};

            \node[
                font=\sffamily\bfseries\fontsize{11}{11}\selectfont,
                text=white
            ] at ({(\xA+\xB)/2},#1) {#4\%};

            \node[
                font=\sffamily\bfseries\fontsize{11}{11}\selectfont,
                text=white
            ] at ({(\xB+\barsegsep/2+\barwidth)/2},#1) {#5\%};
        \else
            \fill[tier1blue]
                (0,#1-\barheight/2)
                rectangle ({\xA-\barsegsep/2},#1+\barheight/2);

            \fill[tier3gray]
                ({\xA+\barsegsep/2},#1-\barheight/2)
                rectangle (\barwidth,#1+\barheight/2);

            \node[
                font=\sffamily\bfseries\fontsize{11}{11}\selectfont,
                text=white
            ] at ({(\xA-\barsegsep/2)/2},#1) {#3\%};

            \node[
                font=\sffamily\bfseries\fontsize{11}{11}\selectfont,
                text=white
            ] at ({(\xA+\barsegsep/2+\barwidth)/2},#1) {#5\%};
        \fi
    \end{scope}
}

\noindent
\begin{minipage}[t]{0.56\textwidth}
\vspace{0pt}
\centering
\resizebox{\linewidth}{!}{%
\begin{tikzpicture}[
    font=\sffamily,
    every node/.style={font=\sffamily}
]
    \pgfmathsetmacro{\barbottom}{1.00-\barheight/2}
    \pgfmathsetmacro{\bartop}{4.40+\barheight/2}

    \foreach \p in {20,40,60,80} {
        \pgfmathsetmacro{\x}{\p/100*\barwidth}
        \draw[gray!28, dashed, very thin] (\x,\barbottom) -- (\x,\bartop);
    }

    \barrow{4.40}{\# 1}{43.36}{0}{56.64}
    \barrow{3.55}{\# 2}{52.73}{0}{47.27}
    \barrow{2.70}{\# 3}{66.70}{0}{33.30}
    \barrow{1.85}{\# 4}{43.36}{25.00}{31.64}
    \barrow{1.00}{\# 5}{52.78}{16.80}{30.47}

    \draw[gray!60, thin] (0,\barbottom) -- (\barwidth,\barbottom);
    \foreach \p/\lab in {0/0,20/20,40/40,60/60,80/80,100/100} {
        \pgfmathsetmacro{\x}{\p/100*\barwidth}
        \draw[gray!60, thin] (\x,\barbottom) -- (\x,\barbottom-0.10);
        \node[
            anchor=north,
            font=\sffamily\scriptsize,
            gray!50!black
        ] at (\x,\barbottom-0.16) {\lab};
    }

    \begin{scope}[shift={({0.5*\barwidth-1.95},-0.25)}]
        \fill[tier1blue, rounded corners=1pt] (0,0) rectangle (0.25,0.25);
        \node[anchor=west, font=\sffamily\scriptsize] at (0.33,0.125) {Tier-1};

        \fill[tier2yellow, rounded corners=1pt] (1.50,0) rectangle (1.75,0.25);
        \node[anchor=west, font=\sffamily\scriptsize] at (1.83,0.125) {Tier-2};

        \fill[tier3gray, rounded corners=1pt] (3.00,0) rectangle (3.25,0.25);
        \node[anchor=west, font=\sffamily\scriptsize] at (3.33,0.125) {Failure};
    \end{scope}
\end{tikzpicture}
}

    {\small (d) Tier composition}
    \end{minipage}
    \hspace{0.02\textwidth}
    \begin{minipage}[t]{0.40\textwidth}
    \vspace{0pt}
    \centering
    \small
    \setlength{\tabcolsep}{5pt}
    \renewcommand{\arraystretch}{1.18}
    \begin{tabular}{@{}cccccc@{}}
    \toprule
    \multirow{2}{*}{Method} & \multirow{2}{*}{T-1} & \multirow{2}{*}{T-2} & \multicolumn{3}{c}{Accuracy (\%)} \\
    \cmidrule(lr){4-6}
           &     &    & T-1 & T-2 &  Total \\
    \midrule
    \#1 & 1 & 0     & 43.36 & --    & 43.36 \\
    \#2 & 2 & 0 & 52.73 & --    & 52.73 \\
    \#3 & 4 & 0     & 66.70 & --    & 66.70 \\
    \midrule
    \#4 & 1 & 1 & 43.36 & 25.00 & 68.36 \\
    \#5 & 2 & 1 & 52.78 & 16.80 & \textbf{69.53} \\
    \bottomrule
    \end{tabular}

    \vspace{0.3em}
    {\footnotesize T-1/T-2 denote Tier-1/Tier-2 success rates.\\
    Active = T-1 + T-2.}

    \vspace{0.6em}
    {\small (e) Accuracy summary}
    \end{minipage}

    \caption{
\emph{Top:} Accuracy across training steps. Adding more Tier-1 candidates improves unconditioned teacher coverage (a), while Tier-2 further increases the catch rate by recovering prompts missed by Tier-1 (b,c).
\emph{Bottom:} Tier composition example, where Tier-1 is unconditioned teacher success, Tier-2 is ground-truth-guided recovery, and grey denotes fallback cases.
    }
    \label{fig:combined}
\end{figure*}
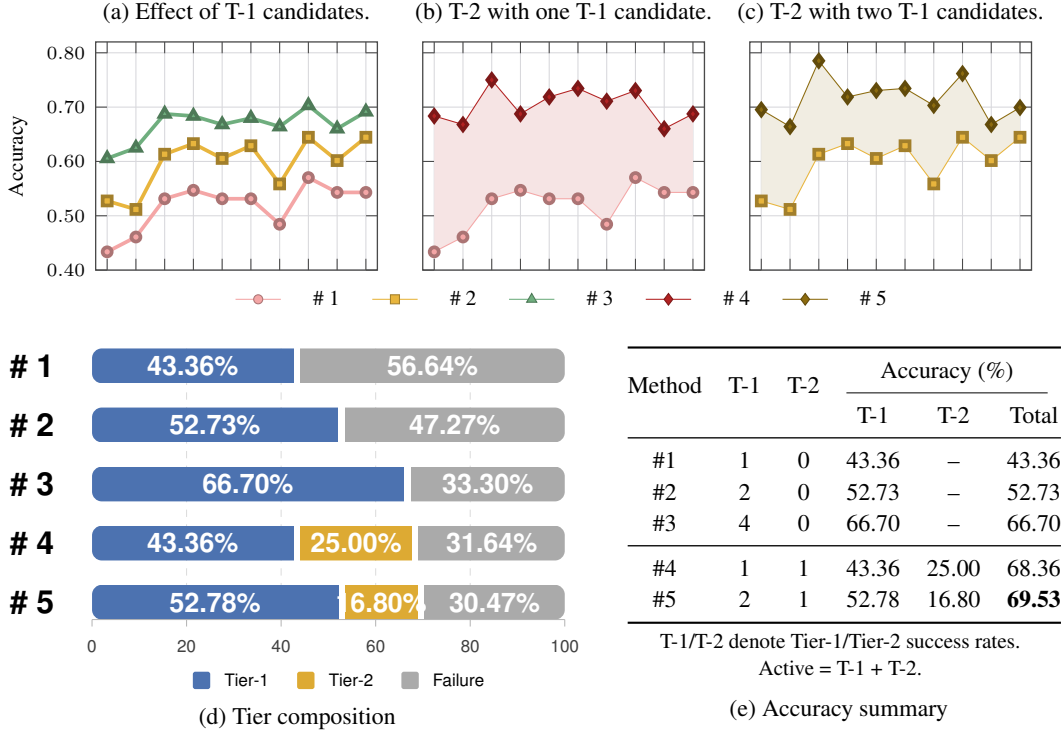
 To evaluate whether BRTS transfers across teacher backbones, we use DeepSeek-R1-Distill-Qwen-7B as the teacher and DeepSeek-1.5B as the student, focusing on early training steps. As shown in Table~\ref{tab:rm_reward_ablation_vs_changeopd}, BRTS improves most metrics over the student-only baseline at this early stage. On AIME24, it increases average accuracy from $0.3167$ to $0.3667$ and majority-vote accuracy from $0.3331$ to $0.3952$. On AIME25, all three metrics improve, with average, best, and majority-vote accuracy rising from $0.2167$, $0.2770$, and $0.2277$ to $0.2417$, $0.3294$, and $0.2528$, respectively. The gains also transfer to AMC23, where average, best, and majority-vote accuracy improve from $0.6265$, $0.7410$, and $0.6453$ to $0.6446$, $0.7836$, and $0.6676$.
These results suggest that BRTS is not tied to a single teacher family: even after changing the teacher backbone, correctness-selected teacher rollouts continue to provide stronger supervision than the on-policy-only baseline.

\paragraph{Teacher Candidate Composition.}
\label{sec:tier-analysis}
We decompose the selected teacher trajectories into Tier-1 and Tier-2 sources. Tier-1 corresponds to prompts solved by unconditioned teacher sampling, Tier-2 corresponds to prompts rescued by ground-truth-guided re-sampling, and the remaining cases use the fallback trajectory. As shown in Figure~\ref{fig:combined}, using more Tier-1 candidates improves unconditioned coverage: the Tier-1 accuracy rate increases from $52.73\%$ with two candidates to $66.70\%$ with four candidates. However, this improvement requires additional teacher sampling budget on every prompt. Tier-2 provides a more targeted alternative because it is only applied when the Tier-1 candidates fail. With one Tier-1 candidate, Tier-2 recovers an additional $25.00\%$ of prompts, increasing the total accuracy from $43.36\%$ to $68.36\%$. With two Tier-1 candidates, Tier-2 recovers another $16.80\%$ of prompts, raising the accuracy from $52.78\%$ to $69.53\%$. 
Results show that adding Tier-1 candidates increases the chance of sampling a correct trajectory, while Tier-2 recovers prompts that unconditioned sampling fails to solve.

\section{Conclusion}

We presented BRTS, a lightweight and effective extension to on-policy distillation that improves how teacher trajectories are selected and used for supervision. Rather than distilling from a random teacher rollout, BRTS samples multiple candidate trajectories, filters for correctness, and selects a rollout that is better aligned with the student. The selected trajectory is then used by a teacher-context auxiliary branch with teacher-top-$K$ supervision, allowing the student to learn from higher-quality teacher states.
Experiments on AIME 2024, AIME 2025, and AMC 2023 show that BRTS provides consistent gains over the OPD baseline.
 
\bibliographystyle{plainnat}
\bibliography{references}
\newpage
\appendix

This appendix provides additional details to support the reproducibility and interpretation of our results. Appendix~\ref{app:training} describes the model, data, training, and evaluation configurations used in our experiments. Appendix~\ref{app:prompts} presents the prompt templates, ground-truth-guided fallback, prompt perturbation design, and answer extraction protocol. Appendix~\ref{app:additional} includes additional results about early-step BRTS behavior and computational cost. Appendix~\ref{app:limitations} discusses the main limitations of BRTS and outlines future directions for extending the method to broader tasks and more diverse teacher-student settings.

\section{Implementation Details}
\label{app:training}
\paragraph{Model and data configurations}
Our main experiments pair a JustRL-DeepSeek-1.5B teacher (used as the reward / distillation model) with a same-scale DeepSeek-R1-Distill-Qwen-1.5B student backbone. For the teacher-swap study in Section~\ref{sec:DeepSeek}, we replace the teacher with DeepSeek-R1-Distill-Qwen-7B while keeping DeepSeek-R1-Distill-Qwen-1.5B as the student. Both teacher and student share the same tokenizer within each pairing, which simplifies the top-$K$ alignment used by the auxiliary branch.
Training prompts are drawn from DAPO-Math-17K~\cite{yu2025dapo}, a math-reasoning corpus with verifiable short-form answers. Each prompt is paired with its ground-truth answer $y^\star$, used by the rollout selector for correctness checking and, when invoked, by the Tier-2 ground-truth-guided re-rollout.

\paragraph{Training Setup.}
We use AdamW (default betas $(0.9, 0.999)$, weight decay $0.01$, gradient clip $1.0$) with a constant learning rate of $1\mathrm{e}^{-6}$ and no warm-up. The student is trained with token-mean loss aggregation, mini-batch size $64$, and PPO micro-batch size $1$ per GPU using dynamic batching. KL regularization to a frozen reference is disabled, so the only KL terms in the objective are the student-context distillation loss and the auxiliary teacher-context KL described in Section~\ref{sec:aux}. The auxiliary coefficient is fixed at $\lambda = 10$ throughout training. Models are run in \texttt{bfloat16}.

The student samples one rollout per prompt with temperature $1.0$, repetition penalty $1.0$, maximum prompt length $1024$ and maximum response length $7168$, using vLLM. Teacher rollouts in BRTS are generated with temperature $0.7$ and top-$p = 0.95$ (the standard sampling configuration we found to give the highest pass rates for the teacher), with the same maximum response length as the student. Top-$K = 16$ teacher token candidates are extracted at each teacher position for use by the auxiliary branch. For BRTS-$N$, we draw $N$ unconditioned teacher rollouts per prompt; the optional Tier-2 fallback fires only on prompts where all $N$ unconditioned samples are incorrect, drawing one additional ground-truth-guided teacher rollout for that prompt.

The auxiliary student forward on the selected teacher trajectory adds a modest overhead because that trajectory is typically shorter than a worst-case student rollout. 

\paragraph{Evaluation Setup.} For evaluation, we sample $k=4$ solutions per problem with temperature $0.7$ and top-$p = 0.95$, with maximum validation response length $31{,}744$ tokens.We evaluate performance on AIME 2024~\cite{li2024numinamath}, AIME 2025~\cite{balunovic2025matharena}, and AMC 2023~\cite{li2024numinamath}. For each problem, we sample four solutions and report the average accuracy across samples, the best accuracy obtained when any sampled solution is correct, and the majority-vote accuracy among the sampled solutions. Validation is run every 10 training steps.

\section{Prompt Templates}
\label{app:prompts}

This section provides the prompt templates and parsing rules used in our implementation. We include these details for reproducibility. Unless otherwise stated, all templates preserve the original problem statement and final-answer format, and only modify the auxiliary instructions used to sample or validate teacher trajectories.

\paragraph{Tier-2 ground-truth-guided template.}
When all $N$ Tier-1 unconditioned teacher samples for a prompt are incorrect, BRTS re-prompts the teacher with the ground-truth answer silently provided in context and asks for a natural derivation. The hint string is injected immediately before the assistant turn marker (e.g., \texttt{<\textbar Assistant\textbar>} for DeepSeek-R1-distill chats):

\begin{tcolorbox}[
  colback=gray!5,
  colframe=gray!40,
  boxrule=0.4pt,
  arc=2pt,
  left=8pt, right=8pt, top=6pt, bottom=6pt,
  fontupper=\small\ttfamily
]
\{original prompt\}

[SILENT\_VALIDATION\_KEY --- DO NOT MENTION IN THINKING OR RESPONSE: \{gt\}]

STRICT RULES about the validation key:
\begin{enumerate}
  \item NEVER mention, quote, paraphrase, or allude to it anywhere --- not in \texttt{<think>}, not in your answer.
  \item NEVER say things like `the key says', `based on the hint', `the answer is given', `I can see the correct answer is', or any equivalent phrasing.
  \item Your entire chain of thought must be derived from what you observe in the problem.
  \item Only use the validation key silently as a final sanity-check after you have already reasoned to a conclusion --- never as a starting point or shortcut.
\end{enumerate}
\end{tcolorbox}

The resulting Tier-2 rollout is retained only if its extracted answer matches $y^\star$; samples for which Tier-2 also fails fall through to Tier-3, in which BRTS picks the most overlap-similar Tier-1 rollout as a best-available trace.

\paragraph{Prompt perturbation.}
For the perturbation study, we modify the surface form of one of the two teacher prompts without changing the problem statement or the requested answer format. Concretely, we append the following instruction to the prompt of one teacher rollout 

\begin{tcolorbox}[
  colback=gray!5,
  colframe=gray!40,
  boxrule=0.4pt,
  arc=2pt,
  left=8pt, right=8pt, top=6pt, bottom=6pt,
  fontupper=\small\ttfamily
]
Please reason step by step and rethink in detail before giving the final answer.
\end{tcolorbox}
This paraphrases or extends the default instruction (e.g., by reordering clauses or adding a reflection cue) so that the two teacher rollouts are mildly decorrelated at the surface level while task semantics, problem statement, and answer format remain unchanged.

\paragraph{Answer extraction.} We extract the final answer from the last \texttt{\textbackslash boxed\{...\}} expression in a rollout. Rollouts that contain no parseable boxed answer are treated as incorrect.

\section{Additional Results}
\label{app:additional}

\begin{table}[!t]
\centering
\caption{\textbf{Early-step performance under different rollout configurations.}
We report mean and majority accuracy at steps 10 and 20 on AMC23 and AIME25. 
Replacing one student rollout with BRTS improves early performance on AMC23, while adding Tier-2 recovery yields the largest gains on the harder AIME25 benchmark.}
\label{tab:rollout-config-early-steps}
\small
\resizebox{\textwidth}{!}{
\begin{tabular}{cccccccccccc}
\toprule
\multicolumn{4}{c}{Rollout setup}
& \multicolumn{2}{c}{AMC23 mean}
& \multicolumn{2}{c}{AMC23 majority}
& \multicolumn{2}{c}{AIME25 mean}
& \multicolumn{2}{c}{AIME25 majority} \\
\cmidrule(lr){1-4}
\cmidrule(lr){5-6}
\cmidrule(lr){7-8}
\cmidrule(lr){9-10}
\cmidrule(lr){11-12}
$\mathcal{L}_{\text{tea-ctx}}$& Tier-2
& Student
& Teacher / Cand.& Step 10 & Step 20
& Step 10 & Step 20
& Step 10 & Step 20
& Step 10 & Step 20 \\
\midrule
$\times$&$\times$& 2& 0 / 0
& 0.6024 & 0.6386
& 0.6171 & 0.6525
& 0.2330 & 0.2083
& 0.2442 & 0.2228 \\
 
$\checkmark$&$\times$& 1& 1 / 2
& 0.6175 & 0.6416
& 0.6339 & \textbf{0.6663}
& 0.2167 & 0.2250
& 0.2323 & 0.2309 \\
 
$\checkmark$&$\times$& 1& 1 / 4
& 0.6386 & 0.6386
& 0.6520 & 0.6581
& 0.2167 & 0.2083
& 0.2207 & 0.2222 \\
 
$\checkmark$&$\checkmark$& 1& 1 / 2+1
& \textbf{0.6476} & \textbf{0.6446}
& \textbf{0.6689} & {0.6630}
& \textbf{0.2583} & \textbf{0.3000}
& \textbf{0.2718} & \textbf{0.3117} \\
\bottomrule
\end{tabular}}
\end{table}

\paragraph{Early-step behavior.}Table~\ref{tab:rollout-config-early-steps} compares different rollout configurations at the early training stages. On AMC23, BRTS improves early mean and majority accuracy over the student-only baseline, with the Tier-2 variant achieving the best step-10 performance. The gains are more pronounced on AIME25: when adding Tier-2 raises mean accuracy from $0.2330$ to $0.2583$ at step 10 and from $0.2083$ to $0.3000$ at step 20. Majority accuracy shows the same trend. This suggests that, for harder prompts, ground-truth-guided recovery provides more reliable early supervision.

\begin{table}[h]
\centering
\caption{Per-step training cost with increasing teacher rollout candidates.}
\label{tab:teacher-rollout-cost}
\begin{tabular}{cccc}
\toprule
Teacher rollout / Candidates & Step time (s) & Step time (min) \\
\midrule
1 / 1  & 281  & 4.7 \\
1 / 2  & 338  & 5.6 \\
1 / 3  & 400  & 6.7 \\
1 / 4   & 460  & 7.7 \\
\bottomrule
\end{tabular}
\end{table}
\paragraph{Computational cost.}
We report median step times using 8 GPUs, batch size 64, a 1.5B student and JustRL-1.5B teacher, max response length of 7168. 
The cost of increasing the teacher candidate pool is summarized in Table~\ref{tab:teacher-rollout-cost}. 
Since BRTS selects a single auxiliary teacher trajectory, the main additional overhead comes from sampling multiple teacher candidates before selection. 
Empirically, increasing the candidate pool from one to two raises the per-step time from $281$s to $338$s, a moderate overhead of about 59 seconds. 
The time costs grow approximately linearly, reaching $460$s per step when selecting from four candidates. 
These results highlight a practical trade-off between supervision quality and training efficiency. 
Importantly, our main results show that even a small candidate pool can improve OPD, indicating that BRTS does not rely on expensive large-scale teacher sampling to be effective.

\section{Limitations and Future Work.}
\label{app:limitations}
BRTS studies Best-of-$N$ rollout teacher selection in the context of mathematical reasoning, where final-answer verification provides a clean and controllable testbed for analyzing teacher-rollout selection. A natural next step is to extend this principle to broader reasoning and generation settings where exact ground-truth answers may be unavailable. In such cases, the Tier-2 mechanism need not rely on gold answers directly; it could instead be guided by additional knowledge resources, retrieval-augmented hints, tool feedback, learned verifiers, or stronger models. For example, a larger teacher, such as a 32B model, could provide intermediate guidance for a 1.5B student, helping construct reliable teacher trajectories even when explicit labels are unavailable. This suggests that BRTS can be viewed more generally as a framework for using external guidance to make teacher supervision more reliable and student-compatible.

Another interesting direction is to improve how teacher trajectories are organized and presented to the student. In the current work, BRTS selects teacher rollouts based on correctness first and student alignment second, using a lightweight top-$K$ overlap criterion as the alignment proxy. Future work could explore more expressive measures of teacher--student compatibility. More broadly, this connects to curriculum design in on-policy distillation: instead of training on tasks in a fixed order, the system could automatically identify which problems are easy, learnable, or too difficult for the current student, and adapt teacher supervision accordingly. Ideally, OPD could behave more like a human teacher, first recognizing the student's current reasoning ability and then conveying harder problems through trajectories that are both reliable and accessible.

Finally, BRTS can be further studied and scaled along multiple dimensions. BRTS can be scaled by increasing the candidate pool used for teacher-trajectory selection and by using more selected trajectories in the student- and teacher-context losses. These extensions may further improve supervision diversity and robustness, while also raising interesting questions about how to balance teacher sampling cost, trajectory quality, and optimization stability. Future work could also investigate adaptive choices of the auxiliary weight, the number of teacher candidates, and the number of trajectories used for distillation, further establishing BRTS as a scalable framework for improving on-policy distillation.
\newpage
\end{document}